\documentclass[lettersize,journal]{IEEEtran}
\usepackage{amsmath,amsfonts}
\usepackage{algorithmic}
\usepackage{algorithm}
\usepackage{array}
\usepackage[caption=false,font=normalsize,labelfont=sf,textfont=sf]{subfig}
\usepackage{textcomp}
\usepackage{stfloats}
\usepackage{url}
\usepackage{verbatim}
\usepackage{graphicx}
\usepackage{cite}
\usepackage{times}
\usepackage{epsfig}
\usepackage{graphicx}
\usepackage{amsmath}
\usepackage{amssymb}
\usepackage{bbm}
\usepackage{booktabs}
\usepackage{xcolor}
\usepackage{array}
\usepackage{multirow}

\newcolumntype{M}{>{\centering\arraybackslash}m{.2\textwidth}}
\newcolumntype{C}[1]{>{\centering\let\newline\\\arraybackslash\hspace{0pt}}p{#1}}
\newcolumntype{R}[1]{>{\raggedleft\let\newline\\\arraybackslash\hspace{0pt}}p{#1}}
\newcolumntype{L}[1]{>{\raggedright\let\newline\\\arraybackslash\hspace{0pt}}p{#1}}

\definecolor{odotred}{RGB}{176,35,24}
\definecolor{odotgreen}{RGB}{78,173,91}
\definecolor{odotblue}{RGB}{45,112,186}
\definecolor{curveorange}{RGB}{233, 105, 24}
\definecolor{curveyellow}{RGB}{255, 224, 63}
\definecolor{curvegreen}{RGB}{5, 212, 97}
\definecolor{curvegray}{RGB}{220, 220, 220}
\definecolor{curveblue}{RGB}{148, 171, 215}

\begin{document}

\title{GSCompleter: A Distillation-Free Plugin for Metric-Aware 3D Gaussian Splatting Completion in Seconds}

\author{Ao Gao, Jingyu Gong$^{\dag}$, Xin Tan, Zhizhong Zhang, Lizhuang Ma, and~Yuan~Xie$^{\ddag}$% <-this % stops a space

\thanks{A. Gao, J. Gong, X. Tan, Z. Zhang, L. Ma, and Y. Xie are with the School of Computer Science and Technology, East China Normal University, Shanghai, China. E-mail: gaoao.cs@stu.ecnu.edu.cn, \{jygong, xtan, zzzhang, yxie\}@cs.ecnu.edu.cn.}% <-this % stops a space

\thanks{A. Gao and Y. Xie are also with the Shanghai Innovation Institute, Shanghai, China.}% <-this % stops a space

\thanks{J. Gong, X. Tan, and Y. Xie are also with Chongqing Key Laboratory of Precision Optics, Chongqing Institute of East China Normal University, Chongqing, China.}% <-this % stops a space

\thanks{J. Gong and Z. Zhang are also with the Shanghai Key Laboratory of Computer Software Evaluating and Testing, Shanghai, China.}% <-this % stops a space

\thanks{L. Ma is also with the Department of Computer Science and Engineering, Shanghai Jiao Tong University, Shanghai, China. E-mail: lzma@sjtu.edu.cn.}% <-this % stops a space

\thanks{$^{\dag}$Corresponding author. $^{\ddag}$Project Leader.}
}

% The paper headers
% \markboth{Journal of \LaTeX\ Class Files,~Vol.~14, No.~8, August~2021}%
% {Shell \MakeLowercase{\textit{et al.}}: A Sample Article Using IEEEtran.cls for IEEE Journals}

% \IEEEpubid{0000--0000/00\$00.00~\copyright~2021 IEEE}
% Remember, if you use this you must call \IEEEpubidadjcol in the second
% column for its text to clear the IEEEpubid mark.

\maketitle

\begin{abstract}
    3D Gaussian Splatting (3DGS) has revolutionized high-fidelity neural rendering with its explicit representation and efficiency. However, reconstructing scenes from sparse viewpoints suffers from severe geometric voids and floaters due to limited coverage. Current scene completion methods typically rely on an iterative ``Repair-then-Distill'' paradigm, which is computationally intensive, prone to unstable optimization, and susceptible to overfitting. To address these limitations, we propose GSCompleter, a distillation-free plugin that shifts scene completion to a stable ``Generate-then-Register'' workflow. Specifically, GSCompleter synthesizes visually plausible 2D reference images and explicitly lifts them into 3D Gaussian primitives with a consistent metric scale via a robust Stereo-Anchor View Selection mechanism. These newly generated primitives are then seamlessly integrated into the global scene using a novel Ray-Constrained Registration strategy. By replacing unstable distillation with rapid geometric registration, GSCompleter exhibits superior 3DGS completion performance across three benchmarks, enhancing both quality and efficiency over various baselines and achieving new state-of-the-art (SOTA) results. Project page: \url{https://yuhuoo.github.io/projects/gscompleter/}.
\end{abstract}


\section{Introduction}
\label{sec:intro}

\begin{figure}
    \centering
    \includegraphics[width=1.0\linewidth]{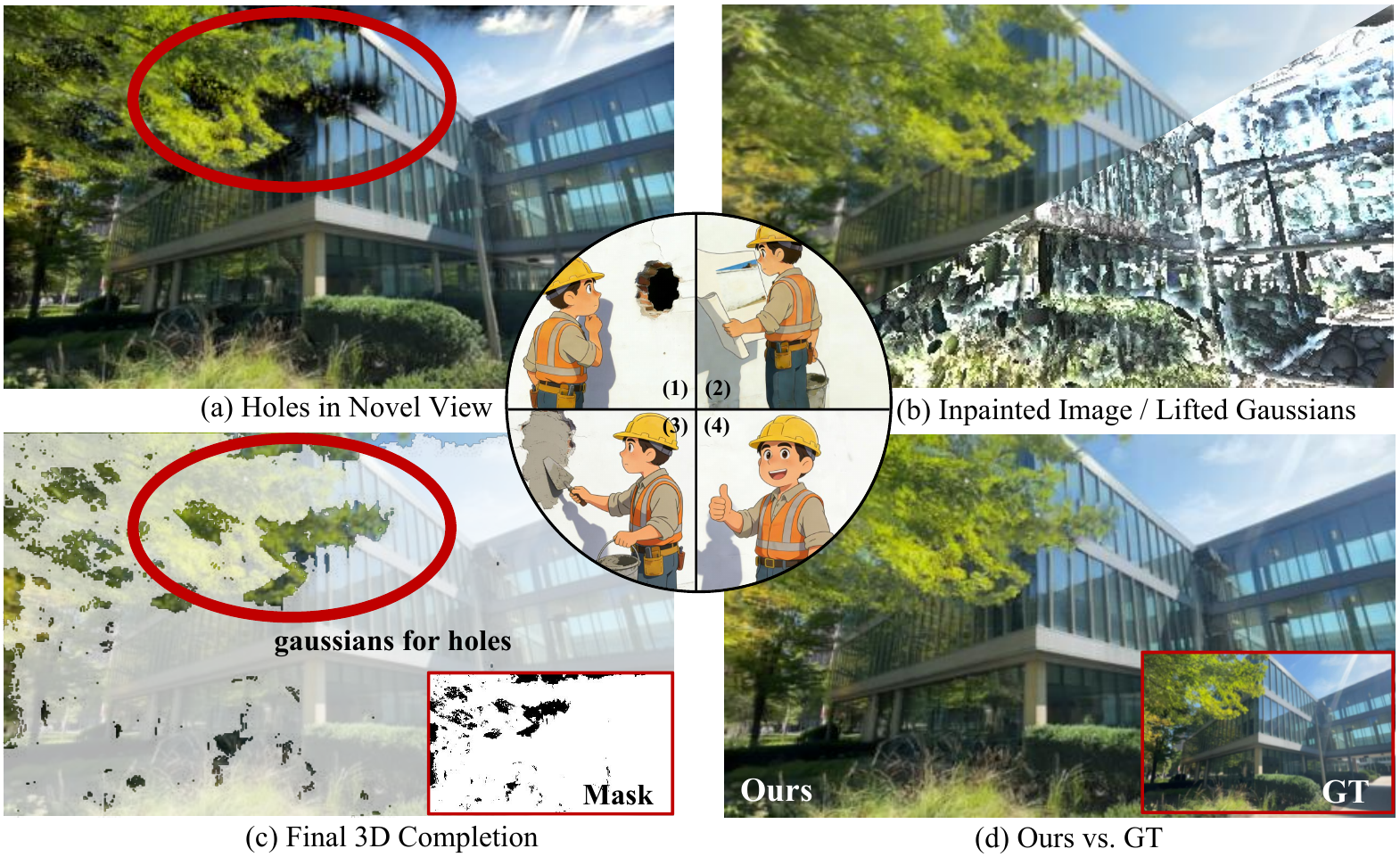}
    \caption{\textbf{Overview of GSCompleter.} We propose a \textbf{``Generate-then-Register''} paradigm for rapid and robust 3DGS scene completion. 
    \textbf{(a)} Given a 3DGS scene exhibiting geometric voids, 
    \textbf{(b)} we first synthesize a high-fidelity 2D reference image via a generative prior and explicitly \textit{lift} it into metric-scale 3D Gaussian primitives guided by a stereo anchor view. 
    \textbf{(c)} Instead of global optimization, we seamlessly register these primitives into the scene via a strictly ray-constrained integration strategy. 
    \textbf{(d)} This process yields a completed scene with fidelity comparable to the Ground Truth, achieved within seconds.}
    \label{fig:intro}
\end{figure}

\IEEEPARstart{N}{eural} rendering has emerged as a cornerstone of 3D reconstruction, autonomous driving, and embodied AI. Among existing techniques, 3D Gaussian Splatting (3DGS)~\cite{kerbl20233dgs} has proven highly effective, offering a superior balance between real-time rendering speed and photo-realistic quality through explicit, differentiable primitives. However, the fidelity of 3DGS remains heavily dependent on the density and coverage of input viewpoints~\cite{li2024dngaussian,xiong2023sparsegs,zhu2024fsgs}. In sparse-view scenarios, 3DGS often fails to reconstruct complete scene geometry, leading to severe artifacts and prominent geometric voids in extrapolated views~\cite{fan2024instantsplat,xu2024mvpgs}.

To address these limitations, state-of-the-art methods typically adopt a ``Repair-then-Distill'' optimization pipeline~\cite{paliwal2025ri3d, liu2024gsenhancer, wu2025difix3d+}. These approaches employ 2D generative priors to inpaint unobserved regions and subsequently distill this information into the 3DGS representation through iterative densification. However, this implicit distillation paradigm suffers from three critical limitations:
\textit{(1) Densification Failure:} In regions lacking initial points, gradient-based optimization struggles to spawn new 3DGS primitives, leading to incomplete geometry and severe floating artifacts.
\textit{(2) Overfitting:} Without explicit 3D supervision, the optimization easily overfits individual views, resulting in floating artifacts near the camera.
\textit{(3) Computational Inefficiency:} Requiring hundreds of optimization steps per viewpoint, the substantial time cost hinders such pipelines from practical deployment.

An alternative is the registration-based approach, which focus on aligning partial Gaussian observations to complete the scene. For instance, GaussReg~\cite{chang2024gaussreg} utilizes an image-guided coarse-to-fine framework to accelerate alignment; however, its heavy reliance on domain-specific training priors limits its generalizability to cross-dataset scenarios. Alternatively, RegGS~\cite{cheng2025reggs} implements a pose-free strategy using Optimal Transport Mixture 2-Wasserstein ($MW_2$) distance to resolve complex $\text{Sim}(3)$ transformations. While it eliminates the need for initial poses, it suffers from significant computational overhead—often requiring minutes to converge—and remains susceptible to local optima. Furthermore, while these methods demonstrate the feasibility of registration-based completion, their reliance on ground-truth images is fundamentally paradoxical given the lack of visual data in unobserved regions, thereby hindering real-world deployment.

Moreover, current registration frameworks typically treat Gaussian primitives as scale-agnostic point clouds, necessitating costly iterative optimization to resolve scale and pose ambiguities. Consequently, the absence of a reliable metric prior remains a primary bottleneck for efficient scene completion. In contrast, recent feed-forward models leverage stereo-depth estimation to directly predict Gaussian primitives with inherent metric scale~\cite{xu2025depthsplat, chen2024mvsplat, wang2025volsplat}, ensuring precise geometric consistency. This breakthrough inspires a pivotal question: \textit{Can we leverage these metric priors to eliminate expensive scale optimization, thereby transforming scene completion into a rapid registration task?}

In this paper, we rethink the unstable ``Repair-then-Distill'' optimization pipeline, advocating a paradigm shift toward a robust ``Generate-then-Register'' workflow. We introduce \textbf{GSCompleter}, a novel plugin for 3DGS scene completion. Our core insight is that Gaussian primitives with accurate metric scale enable near-instant registration, eliminating the need for exhaustive iterative alignment. We first employ a 3D-aware generative prior (PE-Field)~\cite{bai2025positional} to synthesize a plausible image for a novel view. Rather than distilling this content through unstable optimization, we utilize a feed-forward Gaussian model to lift the 2D image into 3D Gaussian primitives with precise physical scale. By leveraging this metric scale advantage, we transform scene completion into a rapid registration problem. This paradigm effectively circumvents the instability of iterative densification while delivering superior rendering fidelity and inference speed.

Results on multiple benchmarks across three categories of baselines (feed-forward, optimization-based, and registration-based) demonstrate that integrating GSCompleter not only consistently boosts PSNR by up to \textbf{2.41 dB}, but also achieves superior stability over optimization-based methods such as vanilla 3DGS with densification (\textbf{19.39 dB} vs. 16.69 dB). Most notably, it accelerates inference by orders of magnitude (\textbf{seconds} vs. minutes) compared to registration-based approaches like RegGS while delivering significantly higher fidelity (\textbf{29.27 dB} vs. 24.65 dB).

Overall, our contributions are summarized as follows:
\begin{itemize}
    \item We propose a novel ``Generate-then-Register'' paradigm that achieves high-fidelity scene completion within seconds, effectively eliminating the geometric artifacts inherent to iterative densification.
    \item We design a Stereo-Anchor View Selection mechanism that identifies optimal reference views to secure robust metric depth estimation, thereby enhancing registration performance.
    \item We introduce a novel ray-constrained strategy to restrict geometric updates along camera rays, enabling fast and precise registration without texture drifting.
    \item Extensive evaluations confirm that GSCompleter demonstrates strong generalizability across diverse deployments, ranging from general indoor and outdoor scenes to challenging autonomous driving scenarios.
\end{itemize}
\section{Related Work}
\label{sec:related}
\noindent\textbf{Feed-forward 3DGS Model.}
Early NeRF-based methods~\cite{barron2021mipnerf, fridovich2022plenoxels, fridovich2023kplanes, barron2023zipnerf} achieved high reconstruction quality but were often limited by a severe trade-off between rendering speed and accuracy~\cite{muller2022instantngp}. To bridge this gap, 3DGS and its variants emerged~\cite{lu2024scaffold, yu2024mipsplatting, zhang2024pixelgs, kheradmand2024mcmc}. However, these methods rely on time-consuming per-scene optimization, which hinders their deployment in real-time applications~\cite{li2024ggrt}.

To overcome these bottlenecks, research has shifted toward generalizable feed-forward models for single-pass inference~\cite{szymanowicz2025flash3d, ye2024noposplat, jin2024lvsm}. Among them, some methods predict Gaussians at each pixel for scene-level reconstruction. Specifically, PixelSplat~\cite{charatan2024pixelsplat} predicts dense probability distributions to localize Gaussian means, whereas MVSplat~\cite{chen2024mvsplat} builds cost volumes via plane sweeping to provide more precise geometric cues. TranSplat~\cite{zhang2025transplat} further incorporates depth confidence and monocular priors to ensure robust feature matching in challenging non-overlapping regions. Additionally, GGN~\cite{zhang2024ggn} and FreeSplat~\cite{wang2024freesplat} focus on optimizing storage density to reduce memory overhead. Recent works have also extended this paradigm to handle unposed inputs~\cite{jiang2025anysplat} and incorporate depth-guided reconstruction~\cite{xu2025depthsplat}. Most recently, VolSplat~\cite{wang2025volsplat} has introduced a voxel-aligned framework to further enhance structural representation. 

Despite these advances, feed-forward 3DGS models still struggle to hallucinate plausible content for unobserved regions, often resulting in visible geometric holes when the viewpoint shifts significantly.

\noindent\textbf{Generative 3D Reconstruction and Completion.} 
Leveraging generative priors for high-fidelity 3D reconstruction has emerged as a highly promising research direction~\cite{sargent2024zeronvs, gao2024cat3d, wu2025genfusion, chung2023luciddreamer, liu2026reconx, liang2025wonderland}. Early methodologies primarily focused on utilizing 2D diffusion models to mitigate artifacts in sparse-view reconstruction by generating interpolated views as pseudo-ground-truth (pseudo-GT) data~\cite{wu2024reconfusion}. To further enhance view extrapolation capabilities, subsequent approaches incorporated video diffusion models~\cite{yu2024viewcrafter}. While these methods successfully synthesize novel views, they typically operate exclusively in the RGB domain, struggling to maintain accurate 3D structural consistency. To address this limitation, SceneCompleter~\cite{chen2025scenecompleter} introduces a geometry-appearance dual-stream diffusion model for 3D scene completion. However, this method operates exclusively within the point cloud space, limiting its applicability to novel view synthesis.

Recent advances in 3D Gaussian Splatting (3DGS) exploit its differentiability to repair scene artifacts through distillation-based optimization. This paradigm typically integrates diffusion models to guide the refinement of Gaussian primitives~\cite{wu2025difix3d+, liu20243dgs-enhancer, yin2025gsfixer}. For instance, ExploreGS~\cite{kim2025exploregs} introduces an information-gain-driven strategy for virtual camera placement, using video diffusion priors to refine rendered results and guide the fine-tuning of Gaussian primitives. Similarly, FlowR~\cite{fischer2025flowr} employs a multi-view flow matching model to directly map low-quality renderings from sparse reconstructions to high-fidelity outputs expected from dense captures. \textbf{RI3D~\cite{paliwal2025ri3d} further refines this paradigm} by learning to denoise artifact-ridden renders or decoupling the synthesis into visible reconstruction and hallucinated inpainting. However, these methods strictly adhere to a ``Repair-then-Distill'' pipeline, which relies on time-intensive iterative distillation to project 2D priors back into 3D space, and they often remain susceptible to near-camera artifacts and high computational overhead.

In contrast, our approach explicitly lifts 2D generated images into 3DGS through a feed-forward model. By circumventing the instabilities of distillation, our method achieves both robust and efficient scene completion.

\noindent\textbf{Registration-based 3D Reconstruction.} 
Point cloud registration remains a fundamental cornerstone of 3D scene reconstruction. Classical algorithms such as ICP~\cite{besl1992ICP}, Go-ICP~\cite{yang2013goicp}, and Fast Global Registration~\cite{zhou2016fasticp} established the groundwork for geometric alignment. With the rise of deep learning, feature-matching frameworks like GeoTransformer~\cite{qin2022geotransformer} and Deep Global Registration~\cite{choy2020deepicp} have significantly enhanced robustness in noisy environments. While earlier NeRF-based approaches~\cite{goli2023nerf2nerf, chen2023dreg} attempt global mapping by aligning local regions, they often suffer from high computational latency and limited generalization due to the heavy overhead of neural fields. Conversely, the discrete nature of 3DGS primitives offers a more flexible representation for alignment~\cite{chang2024gaussreg}. Although recent 3DGS-based registration methods~\cite{cheng2025reggs} show promise in pose-free scenarios, they typically rely on unscaled primitives, necessitating complex $\text{Sim}(3)$ optimizations. Among concurrent developments, VideoLifter~\cite{cong2025videolifter} builds global representations through segmented window matching, while LoopSplat~\cite{zhu2025loopsplat} ensures global consistency in SLAM via pose graph optimization driven by 3DGS registration. The most closely related work to ours is FlexWorld~\cite{chen2025flexworld}, which synthesizes novel views under large pose variations and progressively expands the scene through geometry-aware fusion. However, FlexWorld performs integration within the point cloud space, which limits its overall computational efficiency. 

Unlike these works, we introduce a generative registration plugin that seamlessly integrates with existing feed-forward 3DGS methods for rapid scene completion. By executing registration directly within the Gaussian space, thereby eliminating intermediate representations, our method achieves robust completion in unobserved regions without compromising efficiency.
\section{Method}
\label{sec:methods}
\subsection{Preliminary}

\noindent\textbf{3D Gaussian Splatting (3DGS).} We represent the scene using 3DGS~\cite{kerbl20233dgs}, a set of explicit, differentiable primitives. Each Gaussian is parameterized by its mean position $\boldsymbol{\mu} \in \mathbb{R}^3$, covariance matrix $\mathbf{\Sigma}$, opacity $\alpha \in [0, 1]$, and spherical harmonic (SH) coefficients $\mathbf{c}$. To ensure positive semi-definiteness, the covariance $\mathbf{\Sigma}$ is decomposed into a scaling vector $\mathbf{s} \in \mathbb{R}^3$ and a rotation quaternion $\mathbf{q} \in \mathbb{R}^4$, derived as:
\begin{equation}
\mathbf{\Sigma} = \mathbf{R}\mathbf{S}\mathbf{S}^\top\mathbf{R}^\top,
\end{equation}
where $\mathbf{S} = \text{diag}(\mathbf{s})$ and $\mathbf{R}$ is the rotation matrix from $\mathbf{q}$.
To render these Gaussians into a 2D image, the 3D covariance matrix $\mathbf{\Sigma}$ is projected onto the image plane. Given a viewing transformation $\mathbf{W}$ and the Jacobian of the affine approximation of the projective transformation $\mathbf{J}$, the 2D covariance matrix $\mathbf{\Sigma}'$ in image coordinates is computed as:
\begin{equation}
\mathbf{\Sigma}' = \mathbf{J}\mathbf{W}\mathbf{\Sigma}\mathbf{W}^\top\mathbf{J}^\top.
\end{equation}
For each pixel, the color $C$ is computed by blending $N$ depth-ordered Gaussians overlapping the pixel:
\begin{equation}
C = \sum_{i=1}^{N} \mathbf{c}_i \sigma_i \prod_{j=1}^{i-1} (1 - \sigma_j)
\end{equation}
where $\sigma_i$ is the opacity of the $i$-th Gaussian multiplied by its 2D probability density at the pixel's location. This differentiable rendering pipeline allows for efficient backpropagation to optimize all Gaussian parameters.

\noindent\textbf{Positional-Encoding Field (PE-Field).} We leverage the PE-Field~\cite{bai2025positional} to enable geometry-aware view synthesis. This framework derives dense 3D coordinates $(x, y, z)$ by back-projecting source tokens using monocular depth and re-projecting them into the target frustum. To accommodate both global structure and sub-patch geometric nuances, a hierarchical allocation strategy is implemented to modulate the attention mechanism via 3D-aware Rotary Positional Encodings (RoPE). Specifically, the query vector $\mathbf{Q}^{(h)}$ is partitioned into axial subspaces modulated by level-specific frequencies $l_h$, where higher levels correspond to finer spatial grids:
\begin{equation}
\begin{split}
    \mathbf{Q}^{(h)} &= [\mathrm{RoPE}_{x}^{(l_h)}(\mathbf{Q}_{x}^{(h)}), \mathrm{RoPE}_{y}^{(l_h)}(\mathbf{Q}_{y}^{(h)}), \\
    &\quad \: \mathrm{RoPE}_{z}^{(l_h)}(\mathbf{Q}_{z}^{(h)})].
\end{split}
\end{equation}
By explicitly modeling depth and reasoning over volumetric correspondences, this formulation effectively mitigates depth ambiguity, enforces strict perspective consistency, and ensures that the synthesized content adheres to the underlying 3D scene manifold.

\begin{figure*}[t!]
    \centering
    \includegraphics[width=1.0\linewidth]{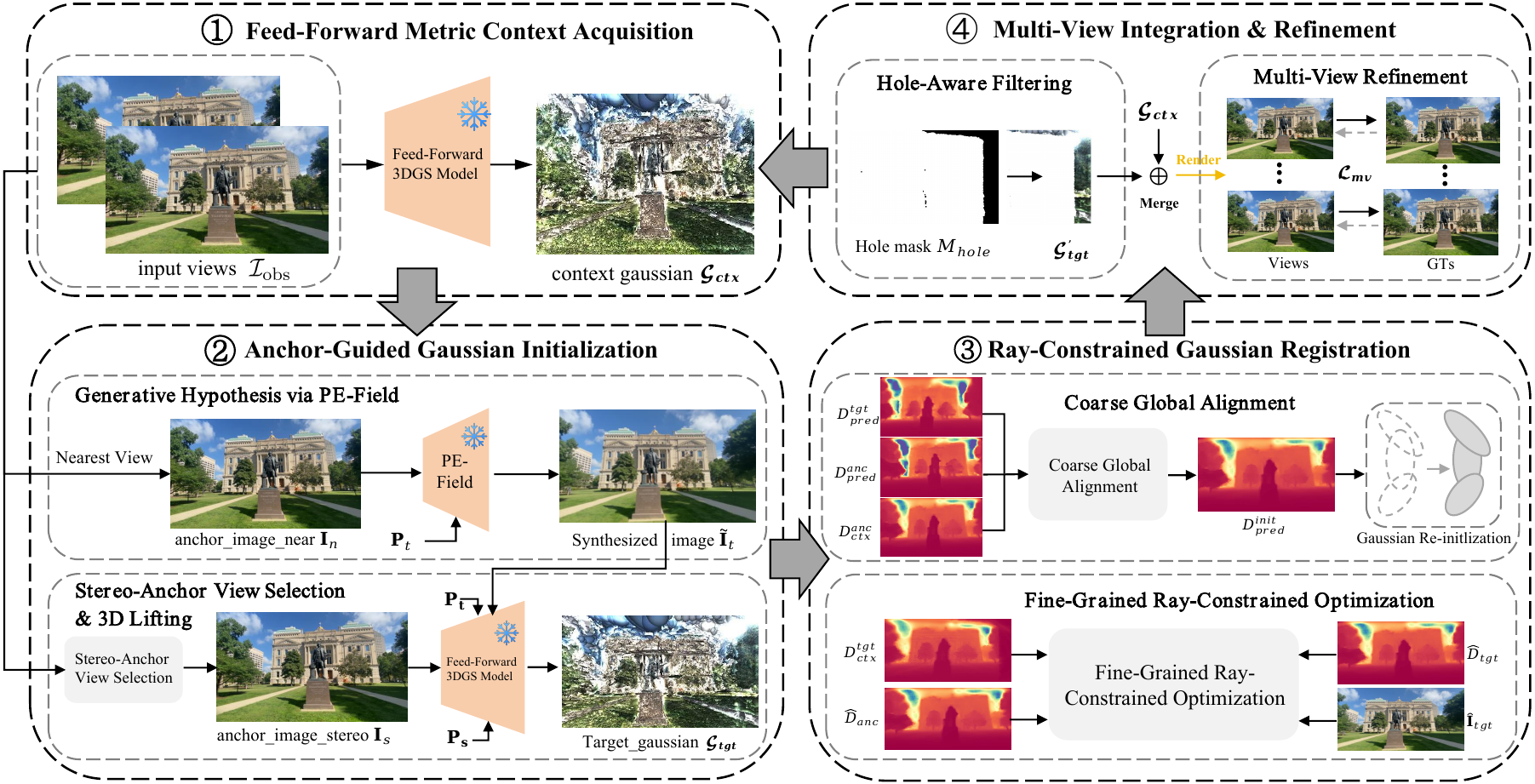}
    \caption{\textbf{Overview of the GSCompleter.} Addressing the geometric holes in the novel view, we adopt a ``Generate-then-Register'' paradigm to complete the scene via four stages: 
    (1) \textbf{Feed-Forward Metric Context Initialization}: We first reconstruct the observed regions using a scale-aware feed-forward 3DGS model, establishing a foundational context with metric scale; 
    (2) \textbf{Anchor-Guided Gaussian Initialization}: To fill the voids, we generatively synthesize the novel view in 2D space, subsequently employing a Stereo-Anchor View Selection mechanism to pair the view with an optimal stereo anchor view, enabling it to be lifted into 3D Gaussians with accurate depth; 
    (3) \textbf{Ray-Constrained Gaussian Registration}: To align these new primitives, we apply a coarse-to-fine mechanism that first rectifies global drift via RANSAC, followed by a strict 1-DoF ray-space optimization to lock primitives along their camera rays for local refinement;
    and (4) \textbf{Multi-View Gaussian Integration \& Refinement}: Finally, redundant primitives are pruned, followed by an opacity-only refinement to seamlessly integrate newly generated Gaussians while preventing catastrophic forgetting of the initial scene.}
    \label{fig:pipeline}
\end{figure*}

\subsection{Overview}
We propose \textbf{GSCompleter}, a novel plugin that shifts the 3DGS completion task from the unstable ``Repair-then-Distill'' optimization to a robust and rapid ``Generate-then-Register'' paradigm. Given a context scene $\mathcal{G}_{\text{ctx}}$ reconstructed from sparse observations $\mathcal{I}_{\text{obs}}$, our goal is to seamlessly integrate missing geometry from a target viewpoint $\mathbf{P}_t$. As illustrated in Fig.~\ref{fig:pipeline}, the proposed pipeline proceeds in four stages: 

\noindent\textbf{(1) Feed-Forward Metric Context Initialization:} We first reconstruct the global context $\mathcal{G}_{\text{ctx}}$ using a pre-trained feed-forward 3DGS model $\Psi(\cdot; \Theta)$, such that $\mathcal{G}_{\text{ctx}} = \Psi(\mathcal{I}_{\text{obs}}; \Theta)$. By leveraging stereo matching priors to establish an absolute metric scale, this initialization effectively reduces the completion problem to a direct registration task.

\noindent\textbf{(2) Anchor-Guided Gaussian Initialization:} To complete unobserved regions, we synthesize a high-fidelity reference image $\tilde{\mathbf{I}}_t$ via the PE-Field generator and \textit{lift} it into a target Gaussian set $\mathcal{G}_{\text{tgt}}$. Crucially, to facilitate reliable metric depth estimation, we introduce a \textit{Stereo-Anchor View Selection} mechanism, which selects optimal context views to establish a robust geometric baseline for subsequent accurate registration.

\noindent\textbf{(3) Ray-Constrained Gaussian Registration:} While the initialization step establishes a metric foundation, minor geometric drifts are inevitable. We address this via a coarse-to-fine registration strategy: we first employ RANSAC to secure a robust global alignment, followed by a novel \textit{1-DoF Ray-Space Optimization}. This step constrains primitives to slide strictly along their camera rays, effectively preventing texture drift. 

\noindent\textbf{(4) Multi-View Gaussian Integration \& Refinement:} Finally, we identify valid completion primitives $\mathcal{G}_{\text{tgt}}'$ by filtering $\mathcal{G}_{\text{tgt}}$ via a hole mask $\mathbf{M}_{\text{hole}}$ and merge them into the global context $\mathcal{G}_{\text{ctx}}$. To prevent catastrophic forgetting during Gaussian integration, we perform an \textit{Opacity-Only Multi-View Refinement}. This process seamlessly fuses the new primitives into the scene while guaranteeing the rendering fidelity of the original views.

\begin{figure}[ht!]
    \centering
    \includegraphics[width=1.0\linewidth]{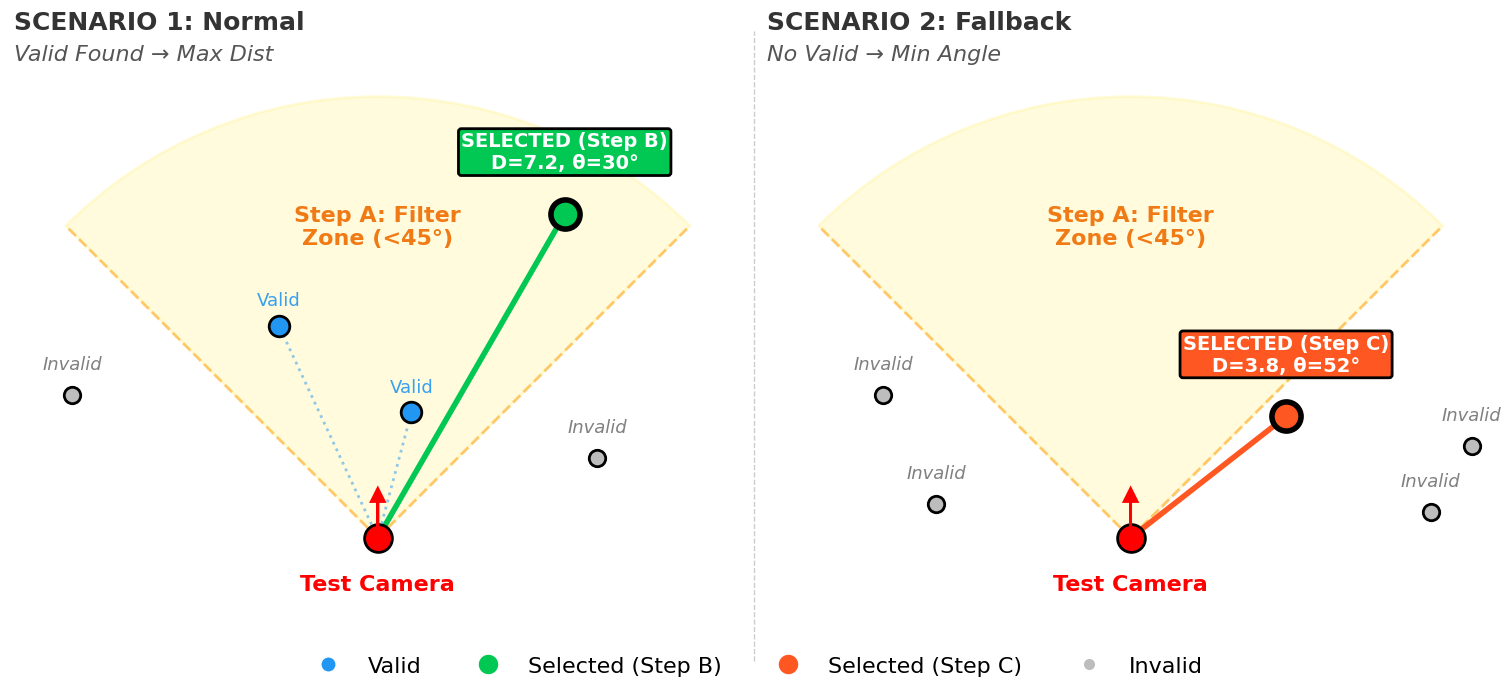}
    \caption{\textbf{Stereo-Anchor View Selection Mechanism.} We identify the optimal reference for 3D lifting through a hierarchical selection strategy: (1) Filtering: Context views with relative rotation $\Delta\theta > 45^\circ$ are discarded to ensure sufficient overlap. (2) Selection: Among valid candidates (Left), we select the one with the maximum baseline to stabilize metric scale. (3) Fallback: In extreme cases where no candidates satisfy the angular constraint (Right), we default to the view with the minimum relative rotation to prevent completion failure.}
    \label{fig:stereo_anchor}
    \vspace{-6pt}
\end{figure}

\subsection{Feed-Forward Metric Context Initialization}
We initiate the pipeline by reconstructing the global context $\mathcal{G}_{\text{ctx}}$ from sparse inputs $\mathcal{V}_{\text{in}} = \{ (\mathbf{I}_i, \mathbf{P}_i) \}_{i=1}^N$ using a pretrained feed-forward 3DGS model $\Psi$:
\begin{equation}
\mathcal{G}_{\text{ctx}} = \Psi(\mathcal{V}_{\text{in}}; \Theta),
\end{equation}
where $\Theta$ denotes the pre-trained 3DGS model parameters. Notably, when employing DepthSplat~\cite{xu2025depthsplat} as our completion baseline, $\Psi$ is instantiated with its pre-trained weights. By incorporating such stereo matching priors, the model guarantees that the output $\mathcal{G}_{\text{ctx}}$ possesses an inherent metric scale for subsequent registration.

\subsection{Anchor-Guided Gaussian Initialization}
With the metric context $\mathcal{G}_{\text{ctx}}$ established, we proceed to hallucinate content for the unobserved target viewpoint $\mathbf{P}_t$. This process involves two coupled steps: 2D hypothesis generation and 3D Gaussians lifting.

\noindent\textbf{Generative Hypothesis via PE-Field}
We synthesize a visual reference image $\tilde{\mathbf{I}}_t$ using the PE-Field ($\Phi$). Notably, we explicitly condition $\Phi$ on the \textit{nearest} observed view $\mathbf{I}_n \in \mathcal{I}_{\text{obs}}$ to inject rich visual context, ensuring the output is factually aligned with the scene:
\begin{equation}
\tilde{\mathbf{I}}_t = \Phi(\mathbf{I}_n, \mathbf{P}_t),
\end{equation}
where $\mathbf{P}_t$ is the target pose. Driven by the underlying 3D positional field, this formulation delivers a trustworthy pseudo-ground truth for the subsequent registration phase.

\noindent\textbf{Stereo-Anchor View Selection \& 3D Lifting.} 
To lift the 2D hypothesis $\tilde{\mathbf{I}}_t$ into 3D Gaussians $\mathcal{G}_{\text{tgt}}$, we face a critical choice regarding Gaussian estimation. While monocular Gaussian estimators like Flash3D~\cite{szymanowicz2025flash3d} or NoPoSplat~\cite{ye2024noposplat} offer rapid inference, they inherently suffer from scale ambiguity, predicting geometry in a canonical space that does not align with the global scene directly. To ensure geometric and feature-space consistency, we employ the same stereo-based feed-forward 3DGS model used in the context acquisition stage. 

Furthermore, rather than adopting the nearest neighbor from the preceding stage, which often degrades multi-view stereo (MVS) depth estimation, we select an optimal \textit{stereo} view to enforce valid geometric constraints. As illustrated in Fig.~\ref{fig:stereo_anchor}, we select the stereo view $\mathbf{P}_s$ based on two criteria: (1) a rotation constraint $\Delta\theta < 45^\circ$ to ensure sufficient feature overlap for correspondence matching; and (2) maximizing baseline distance to minimize triangulation error. Then we construct an image pair $\mathcal{V}_{\text{pair}} = (\tilde{\mathbf{I}}_t, \mathbf{I}_s)$ by coupling the target hypothesis with the selected anchor. This pair is then fed into the regressor:
\begin{equation}
\mathcal{G}_{\text{tgt}} = \Psi(\mathcal{V}_{\text{pair}}; \Theta).
\end{equation}
This strategy offers a key benefit: the newly generated and context Gaussians effectively achieve implicit alignment between their coordinate systems.

\subsection{Ray-Constrained Gaussian Registration}
Although the stereo-guided initialization establishes a solid metric foundation, subtle depth discrepancies remain inevitable since the depth estimations for the context and completed regions are performed independently. To bridge these minor misalignments, we employ a coarse-to-fine registration strategy, ensuring that the newly generated Gaussian primitives are seamlessly integrated into the global scene within a unified coordinate system.

\noindent\textbf{Coarse Global Alignment.} Since the feed-forward model $\Psi$ outputs paired depth maps for both the Anchor View ($D_{\text{pred}}^{\text{anc}}$) and the Target View ($D_{\text{pred}}^{\text{tgt}}$), we utilize the Anchor View as a robust bridge to align the predicted depth $D_{\text{pred}}^{\text{anc}}$ with the established context depth $D_{\text{ctx}}^{\text{anc}}$. This choice is motivated by the limitations of the projected target depth $D_{\text{ctx}}^{\text{tgt}}$, which is inherently sparse and prone to geometric voids. In contrast, the Anchor View provides dense geometric correspondences, enabling a more stable and robust alignment. We formulate the mapping from the predicted depth to the context depth in the anchor view as a global affine transformation and robustly estimate the parameters using RANSAC:
\begin{equation}\min_{s, t} \sum_{\mathbf{p} \in \Omega} \rho \left( \left| D_{\text{ctx}}^{\text{anc}}(\mathbf{p}) - (s \cdot D_{\text{pred}}^{\text{anc}}(\mathbf{p}) + t) \right| \right),
\end{equation}
where $\Omega$ denotes the set of pixels with valid depth values. These parameters are then applied to the Target View to update the target Gaussians ($D_{\text{tgt}}^{\text{init}} = s \cdot D_{\text{pred}}^{\text{tgt}} + t$). It is worth noting that the affine alignment $(s, t)$ is not intended to recover the metric scale from scratch; rather, it serves as a critical fine-tuning step to eliminate systematic residuals. This step ensures the new primitives are geometrically consistent with the global map, providing a valid initialization for the subsequent optimization.

\begin{figure}[t!]
    \centering
    \includegraphics[width=1.0\linewidth]{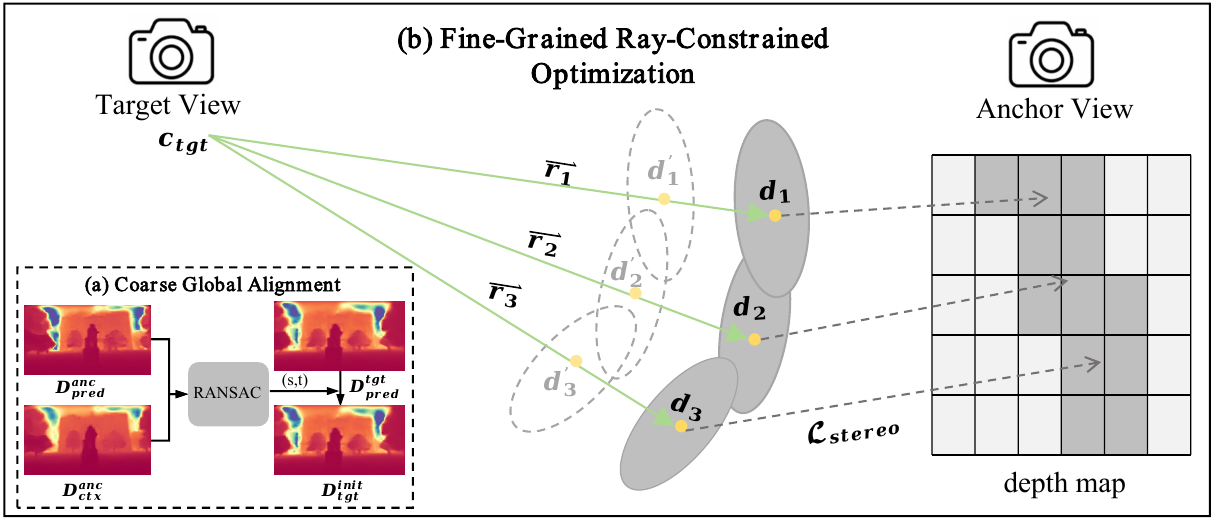}
    \caption{\textbf{Ray-Constrained Gaussian Registration.} 
    (a) \textbf{Coarse Global Alignment}: We employ RANSAC to estimate the global affine parameters ($s, t$), which are used to re-initialize the depth of the target Gaussians. 
    (b) \textbf{Fine-grained Ray-Constrained Optimization}: We optimize the Gaussian depth solely by adjusting the distance along the camera ray. Concurrently, we reproject these primitives into the stereo anchor view to enforce precise geometric structural accuracy.}
    \label{fig:Ray_Constrain}
\end{figure}

\noindent\textbf{Fine-Grained Ray-Constrained Optimization.} 
While rigid alignment corrects the global scale drift, it cannot resolve local non-linear residuals caused by disparity estimation. To correct these geometric distortions without degrading visual fidelity, we propose a 1-DoF Ray-Space Optimization, which consists of two steps: \textbf{(1) Ray-Space Unprojection.} Since $D_{\text{tgt}}^{\text{init}}$ represents the planar Z-depth, we first unproject it into the Euclidean ray space. For each primitive $i$ in the target prediction, we derive the initial Euclidean distance $d_i^{\text{init}}$ along its viewing ray. Formally, we convert the planar depth value sampled at the primitive's projected 2D coordinate $\mathbf{u}_i$ into the ray distance via geometric correction:
\begin{equation}
d_i^{\text{init}} = \frac{D_{\text{tgt}}^{\text{init}}(\mathbf{u}_i)}{\mathbf{r}_i \cdot \mathbf{v}_{\text{view}}},
\end{equation}
where $\mathbf{r}_i$ denotes the normalized ray direction and $\mathbf{v}_{\text{view}}$ represents the camera's principal axis. We initialize the learnable distance parameter $d_i$ with this value: $d_i \leftarrow d_i^{\text{init}}$. \textbf{(2) 1-DoF Optimization.} As illustrated in Fig.~\ref{fig:Ray_Constrain}, we constrain the optimization strictly along the viewing rays. By freezing all non-positional attributes (rotation, scaling, opacity), the 3D position $\boldsymbol{\mu}_i$ of each Gaussian is parameterized solely by its scalar distance $d_i$ along the ray:
\begin{equation}
\boldsymbol{\mu}_i(d_i) = \mathbf{c}_{\text{tgt}} + d_i \cdot \mathbf{r}_i,
\end{equation}
where $\mathbf{c}_{\text{tgt}}$ denotes the camera center of the target view, and $\mathbf{r}_i$ is the normalized direction vector of the viewing ray passing through the $i$-th primitive. With this parameterization, we directly optimize the set of scalar distances $\mathcal{D} = \{d_i\}$ by minimizing the joint objective:
\begin{equation}
\mathcal{L}_{\text{total}} = \lambda_d \mathcal{L}_{\text{depth}} + \lambda_s \mathcal{L}_{\text{stereo}} + \lambda_c \mathcal{L}_{\text{rgb}}.
\end{equation}
Specifically, $\mathcal{L}_{\text{depth}} = \| \hat{D}_{\text{tgt}} - D_{\text{ctx}}^{\text{tgt}} \|_1$ anchors the predicted geometry to the sparse global map ($\lambda_d=1.0$); $\mathcal{L}_{\text{stereo}} = \| \hat{D}_{\text{anc}} - D_{\text{ctx}}^{\text{anc}} \|_1$ enforces multi-view consistency by minimizing the error between the rendered depth and the established anchor depth ($\lambda_s=1.0$); and $\mathcal{L}_{\text{rgb}} = \| \hat{I}_{\text{tgt}} - \tilde{\mathbf{I}}_t \|_1$ acts as a weak photometric regularizer ($\lambda_c=0.1$) to preserve visual fidelity. In this formulation, $\hat{D}_{\text{tgt}}$ and $\hat{I}_{\text{tgt}}$ are rendered from the composite scene (integrating $\mathcal{G}_{\text{ctx}}$ and $\mathcal{G}_{\text{tgt}}$), whereas $\hat{D}_{\text{anc}}$ is rendered from the target Gaussians $\mathcal{G}_{\text{tgt}}$ under the Anchor View.

\subsection{Multi-View Gaussian Integration \& Refinement}

\noindent\textbf{Hole-Aware Filtering \& Integration.} 
Before integration, we filter the aligned target Gaussians $\mathcal{G}_{\text{tgt}}$ to avoid redundancy. We render the opacity map $\mathbf{A}_{\text{ctx}}$ from the context $\mathcal{G}_{\text{ctx}}$ and derive a binary hole mask $\mathbf{M}_{\text{hole}}$ via thresholding:
\begin{equation}
\mathbf{M}_{\text{hole}}(\mathbf{p}) = \mathbbm{1}(\mathbf{A}_{\text{ctx}}(\mathbf{p}) < \tau),
\end{equation}
where $\tau=0.5$ and $\mathbbm{1}(\cdot)$ is the indicator function. This mask identifies unobserved regions in the current map. Consequently, only primitives from $\mathcal{G}_{\text{tgt}}$ that fall within these hole regions are retained. Specifically, for pixel-aligned primitives, we directly select indices based on $\mathbf{M}_{\text{hole}}$; for voxel-aligned primitives, we determine validity by projecting them onto the mask. Finally, the filtered primitives $\mathcal{G}_{\text{tgt}}'$ are merged into the global context: $\mathcal{G}_{\text{ctx}} \leftarrow \mathcal{G}_{\text{ctx}} \cup \mathcal{G}_{\text{tgt}}'$.

\noindent\textbf{Opacity-Only Multi-View Refinement.} 
To prevent catastrophic forgetting of the established map, we perform a multi-view optimization. Specifically, we freeze the existing context $\mathcal{G}_{\text{ctx}}$ and optimize only the opacity $\alpha$ of the new primitives $\mathcal{G}_{\text{tgt}}'$, while freezing other attributes (position, rotation, scaling, and SH). The objective is:
\begin{equation}
\mathcal{L}_{\text{mv}} = \lambda \| \hat{I}_{\text{tgt}} - \mathbf{\tilde{I}}_t \|_1 +\sum_{k \in \mathcal{N}_t} \| \hat{I}_k - I_k \|_1,
\label{eq:loss_mv}
\end{equation}
where $\hat{I}_k$ and $I_k$ represent the rendered and ground-truth images at viewpoint $k$, respectively, and $\mathcal{N}_t$ denotes a set of context views. The first term enforces multi-view consistency, while the second term aligns the target rendering with the pseudo-ground truth. In our experimental setting, we set $k=2$ and $\lambda = 0.1$. This lightweight refinement effectively prevents artifacts in existing views, allowing for a seamless and consistent incremental expansion of the scene.
\section{Experiments}
\label{sec:exp}

\subsection{Datasets and Evaluation Protocol.}
\label{subsec:datasets_evaluation}

\noindent\textbf{Datasets.} Following DepthSplat~\cite{xu2025depthsplat}, we evaluate GSCompleter on RealEstate10K~\cite{zhou2018stereo}, ACID~\cite{liu2021infinite}, and DL3DV~\cite{ling2024dl3dv}. Specifically, the test sets consist of 7,281 scenes for RealEstate10K, 1,972 scenes for ACID, and 140 scenes for DL3DV. We conduct fair comparisons between our method and the baselines across these datasets.

\noindent\textbf{Metrics.} We employ PSNR, SSIM, and LPIPS to evaluate novel view quality. To evaluate geometric consistency, we report AbsRel, Chamfer Distance (CD), and F-Score. These metrics quantify the alignment error between the generated target primitives and the context primitives.

\noindent\textbf{Baselines.} We evaluate GSCompleter by adding it to state-of-the-art models: MVSplat~\cite{chen2024mvsplat}, DepthSplat~\cite{xu2025depthsplat}, and VolSplat~\cite{wang2025volsplat}. As a plugin, GSCompleter is integrated into these baselines (denoted as \textit{Method+Ours}) to show performance gains. Our evaluation primarily focuses on pixel-aligned methods (MVSplat and DepthSplat). To demonstrate generality, we also extend to the voxel-aligned method (VolSplat). Additionally, we compare our approach against the optimization-based baseline (Vanilla 3DGS w/ densification) and the registration-based baseline RegGS~\cite{cheng2025reggs}.

\begin{figure}[t!]
   \centering
   \includegraphics[width=1.0\linewidth]{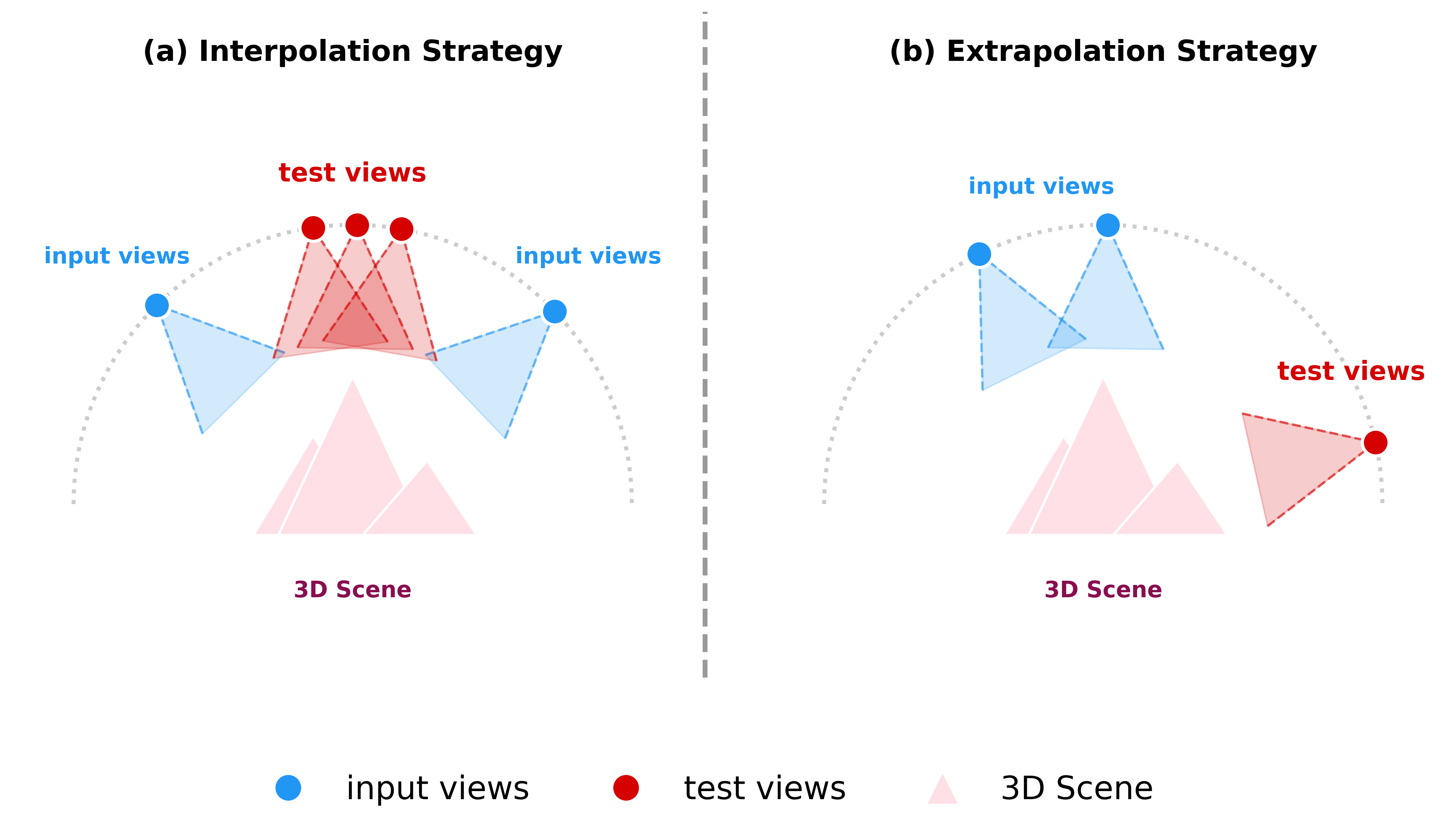}
   \caption{\textbf{Visual Illustration} of the 2-view input and 1-view target extrapolation setting.}
   \label{fig:evaluation_strategy} 
\end{figure}

\noindent\textbf{Evaluation Protocol.} Unlike baselines that primarily focus on view interpolation, we design a 2-view extrapolation setting to validate the model's ability to synthesize large-scale unobserved regions (See Fig.~\ref{fig:evaluation_strategy} for details). We define this evaluation protocol as an $n$-$k$ configuration, where two context views $(I_{ctx}^1, I_{ctx}^2)$ are selected from a sequence of $k$ frames to synthesize a third unobserved target view $I_{tgt}$. Specifically, we employ an $n$-$30$ setting for RealEstate10K and ACID, and an $n$-$10$ setting for DL3DV to account for its motion complexity. Consequently, this setup serves as a rigorous benchmark for geometric and textural completion, shifting the paradigm from simple novel-view interpolation to challenging scene completion.

\subsection{Implementation Details.}
\label{subsec:implementation}

\noindent GSCompleter is implemented in PyTorch and all experiments are conducted on a single NVIDIA H200 GPU. Following DepthSplat~\cite{xu2025depthsplat}, we set images resolutions to $256 \times 256$ for RealEstate10K/ACID and $256 \times 448$ for DL3DV. During inference, we perform 50 iterations of Ray-Constrained Registration ($lr=0.01$), followed by 30 iterations of Opacity-Only Refinement ($lr=0.08$). For the generative prior, we employ only 4 inference steps for the PE-Field, which is sufficient to achieve high-fidelity content. All baseline models maintain frozen pre-trained weights throughout the entire process.

\begin{figure*}[ht!] 
    \centering \includegraphics[width=1.0\linewidth]{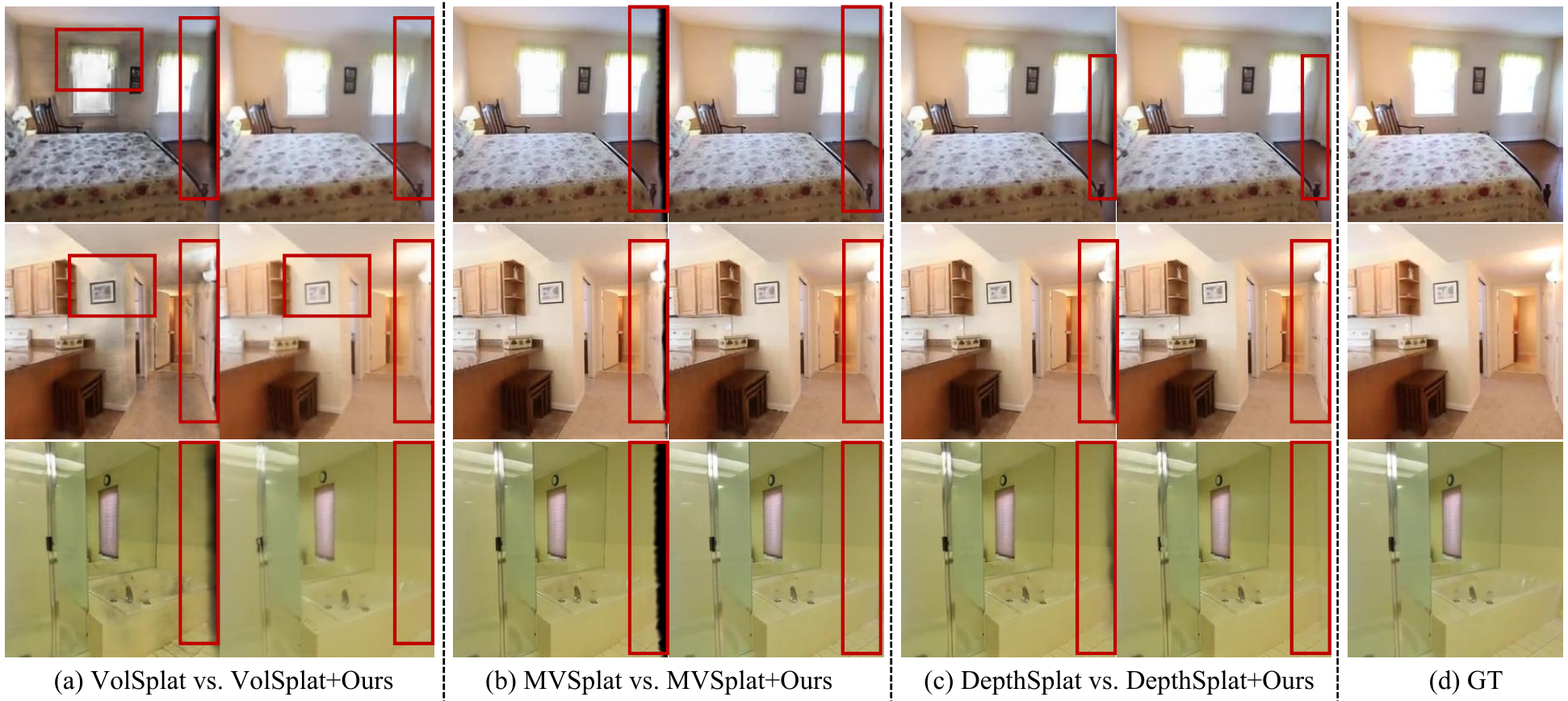}
    \caption{Qualitative Comparison on RealEstate10K. While baselines exhibit significant geometric collapses or ``black holes'' in unobserved regions, our method achieves high fidelity consistent with the Ground Truth (GT). GSCompleter accurately recovers complex geometric structures and scene details while maintaining robustness across diverse baseline architectures (\textit{e.g.}, pixel-aligned and voxel-aligned).}
    \label{fig:qualitative_compare} 
\end{figure*}

\subsection{Experimental Results}
\label{subsec:experimental_results}

\noindent\textbf{Quantitative Results.} 
Tables~\ref{tab:re10k_results}, \ref{tab:acid_results}, and \ref{tab:dl3dv_results} summarize the performance across three benchmarks. GSCompleter consistently outperforms all baselines, achieving state-of-the-art results. For instance, on RealEstate10K, our module yields a significant \textbf{+2.41~dB} PSNR gain for MVSplat.

\noindent\textbf{Qualitative Results.} 
Futhermore, as illustrated in Fig.~\ref{fig:qualitative_compare}, baseline methods exhibit severe geometric voids in unobserved regions due to limited view coverage. By integrating GSCompleter, these black holes are effectively completed with physically plausible geometry and coherent textures, yielding results highly consistent with the Ground Truth (GT). This performance demonstrates the effectiveness of our generative prior in synthesizing realistic scene content. 

\subsection{Analysis of Versatility Across Diverse Architectures}
\label{subsec:Versatility}

We further analysis the plug-and-play versatility of GSCompleter across different feed-forward 3DGS architectures:

\noindent\textbf{(1) Pixel-Aligned Baselines (MVSplat and DepthSplat):} GSCompleter yields comprehensive improvements across all metrics. This performance boost is attributed to our Ray-Constrained Registration, which intrinsically matches the pixel-aligned geometric nature of these methods. By strictly enforcing consistency along camera rays, our paradigm ensures both high-fidelity reconstruction and precise texture alignment.

\noindent\textbf{(2) Voxel-Aligned Baseline (VolSplat):} Our method achieves significant gains by filling geometric voids (e.g., +2.00~dB PSNR on ACID). The slight LPIPS rise stems from the structural conflict between our continuous ray-based registration and VolSplat's discrete voxel grid, where projecting ray-aligned Gaussians into a grid representation introduces minor misalignment.

\begin{table}[ht!]
    \caption{Quantitative comparison on RealEstate10K. Bold denotes the best performance. \textcolor[RGB]{0,150,0}{Green} and \textcolor{red}{Red} indicate performance improvement and degradation relative to each baseline, respectively.}
    \label{tab:re10k_results}
    \centering
    \begin{tabular}{llll}
    \toprule
    Method & PSNR $\uparrow$ & SSIM $\uparrow$ & LPIPS $\downarrow$ \\
    \midrule
    VolSplat & 21.62 & 0.855 & 0.159 \\
    VolSplat+Ours & 23.54~\tiny{\textcolor[RGB]{0,150,0}{(+1.92)}} & 0.860~\tiny{\textcolor[RGB]{0,150,0}{(+0.005)}} & 0.162~\tiny{\textcolor{red}{(+0.003)}} \\
    \midrule
    MVSplat & 24.93 & 0.873 & 0.132 \\
    MVSplat+Ours & 27.34~\tiny{\textcolor[RGB]{0,150,0}{(+2.41)}} & 0.884~\tiny{\textcolor[RGB]{0,150,0}{(+0.011)}} & 0.122~\tiny{\textcolor[RGB]{0,150,0}{(-0.010)}} \\
    \midrule
    DepthSplat & 25.87 & 0.883 & 0.125 \\
    DepthSplat+Ours & \textbf{27.60~\tiny{\textcolor[RGB]{0,150,0}{(+1.73)}}} & \textbf{0.889~\tiny{\textcolor[RGB]{0,150,0}{(+0.006)}}} & \textbf{0.117~\tiny{\textcolor[RGB]{0,150,0}{(-0.008)}}} \\
    \bottomrule
    \end{tabular}
\end{table}

\begin{table}[ht!]
    \caption{Quantitative comparison on ACID dataset. Color coding follows Table~\ref{tab:re10k_results}.}
    \label{tab:acid_results}
    \centering
    \begin{tabular}{llll}
    \toprule
    Method & PSNR $\uparrow$ & SSIM $\uparrow$ & LPIPS $\downarrow$ \\
    \midrule
    VolSplat & 21.39 & 0.810 & 0.208 \\
    VolSplat+Ours & 23.39~\tiny{\textcolor[RGB]{0,150,0}{(+2.00)}} & 0.820~\tiny{\textcolor[RGB]{0,150,0}{(+0.010)}} & 0.209~\tiny{\textcolor{red}{(+0.001)}} \\
    \midrule
    MVSplat & 27.15 & 0.848 & 0.146 \\
    MVSplat+Ours & \textbf{28.60~\tiny{\textcolor[RGB]{0,150,0}{(+1.45)}}} & 0.855~\tiny{\textcolor[RGB]{0,150,0}{(+0.007)}} & 0.139~\tiny{\textcolor[RGB]{0,150,0}{(-0.007)}} \\
    \midrule
    DepthSplat & 27.30 & 0.853 & 0.141 \\
    DepthSplat+Ours & 28.58~\tiny{\textcolor[RGB]{0,150,0}{(+1.28)}} & \textbf{0.857~\tiny{\textcolor[RGB]{0,150,0}{(+0.004)}}} & \textbf{0.134~\tiny{\textcolor[RGB]{0,150,0}{(-0.007)}}} \\
    \bottomrule
    \end{tabular}
\end{table}
\begin{table}[ht!]
    \caption{Quantitative comparison on DL3DV dataset. Color coding follows Table~\ref{tab:re10k_results}.}
    \label{tab:dl3dv_results}
    \centering
    \begin{tabular}{llll}
    \toprule
    Method & PSNR $\uparrow$ & SSIM $\uparrow$ & LPIPS $\downarrow$ \\
    \midrule
    DepthSplat & 21.49 & 0.748 & 0.185 \\
    DepthSplat+Ours & \textbf{22.45~\tiny{\textcolor[RGB]{0,150,0}{(+0.96)}}} & \textbf{0.750~\tiny{\textcolor[RGB]{0,150,0}{(+0.002)}}} & \textbf{0.182~\tiny{\textcolor[RGB]{0,150,0}{(-0.003)}}} \\
    \bottomrule
    \end{tabular}
\end{table}

\subsection{Analysis of Completion Paradigms}
\label{subsec:completion_paradigms_compare}
\begin{figure}[t!]
    \centering
    \includegraphics[width=1.0\linewidth]{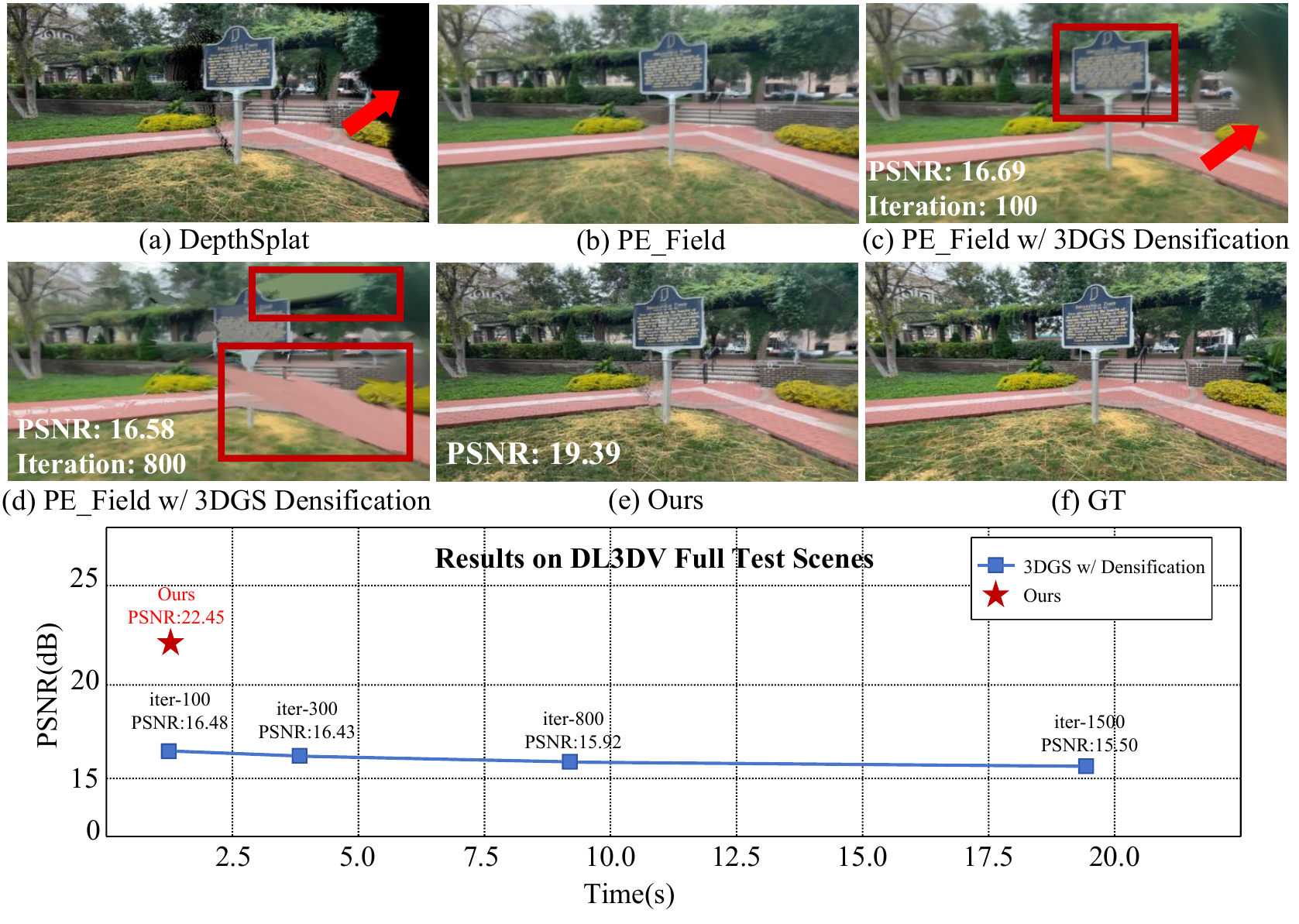}
    \caption{\textbf{Analysis of Completion Paradigms with Optimization-based Baselines.} While densification tends to overfit the reference view, our method effectively mitigates this issue.}
    \label{fig:fig_densification}
\end{figure}

To highlight the advantages of our ``Generate-then-Register'' paradigm, we compare it against two alternative approaches:

\noindent\textbf{vs. Optimization-based Baselines (Fig.~\ref{fig:fig_densification}).}
We adopt the standard densification process of Vanilla 3DGS as our baseline. Naive densification faces a dilemma: insufficient optimization (100 iters) leaves geometric voids, while intensive optimization (800 iters) leads to severe overfitting. This overfitting manifests as obstructive artifacts near the camera, whereas our method ensures robust reconstruction with high fidelity comparable to the ground truth (GT).

\begin{figure}[t!]
    \centering
    \includegraphics[width=1.0\linewidth]{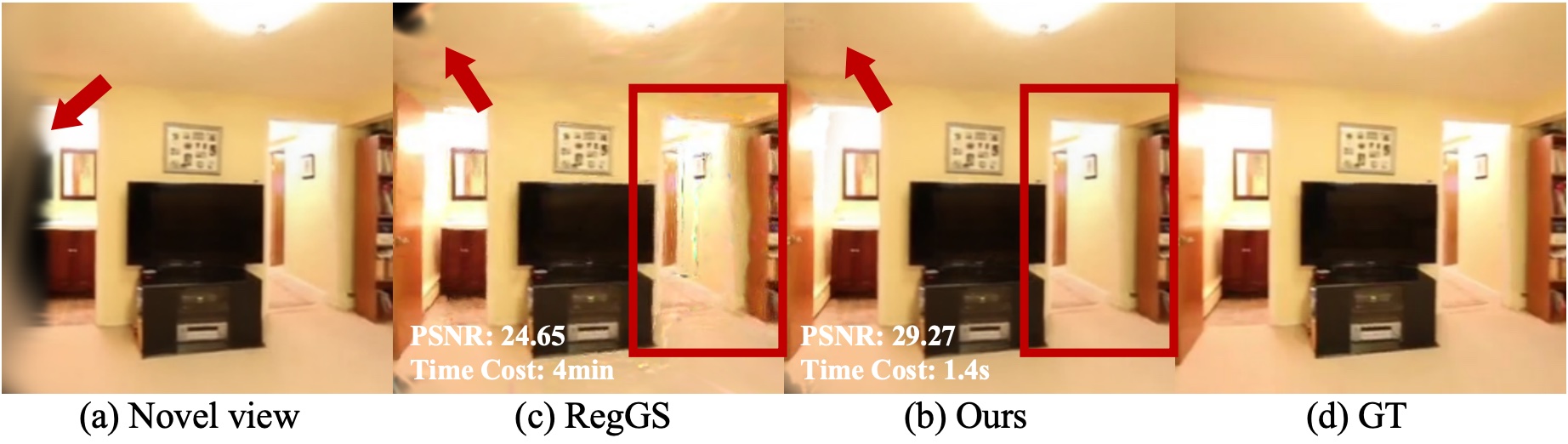}
    \caption{\textbf{Analysis of Completion Paradigms with Registration-based Baselines.} RegGS suffers from severe geometric distortions (highlighted in red) arising from scale-agnostic optimization. In contrast, our method leverages metric priors to achieve precise alignment while strictly preserving structural fidelity. Given the same input views, our geometric pipeline accelerates the process by over \textbf{170$\times$} compared to RegGS (\textbf{1.43s} vs. $\sim$4 min).}
    \label{fig:reggs_compare}
\end{figure}

\noindent\textbf{vs. Registration-based Baselines (Fig.~\ref{fig:reggs_compare}).}
RegGS~\cite{cheng2025reggs} suffers from scale drift and high computational cost (measured in minutes) due to optimizing unscaled primitives via the expensive $MW_2$ distance. In contrast, our approach leverages stereo priors to directly estimate Gaussian primitives with accurate scale, enabling precise and rapid registration in seconds.

\subsection{Ablation Study}
\label{subsec:ablation}

% \noindent\textbf{Geometric Registration Accuracy.} 
Table~\ref{tab:ablation} evaluates the contribution of each core component to visual fidelity and geometric registration accuracy. 

\noindent\textbf{w/o Stereo-Anchor View Selection (SA):} 
Removing SA (\textit{w/o SA}) results in the most significant performance drop, with the Chamfer Distance (CD) increasing to 3.977 and F-Score decreasing to 0.375, confirming that stereo parallax is the primary determinant for resolving scale ambiguity. 

\noindent\textbf{w/o Depth-Alignment (DA):} 
Disabling the global alignment module (\textit{w/o DA}) degrades the F-Score ($0.465 \rightarrow 0.433$), validating that explicit global rectification is essential to correct residual geometric drift. 

\noindent\textbf{w/o Ray-Constrained Optimization (RC):} 
Compared to unconstrained optimization, our RC improves geometric fidelity (CD $3.491 \rightarrow 3.334$) while maintaining high rendering quality by restricting Gaussian movements along camera rays. 

\noindent\textbf{w/o All Components:} 
The results demonstrate that a naive combination of generative content and registration is insufficient for high-quality 3D scene completion. Simply integrating generative outputs without our proposed modules leads to a significant degradation in both geometric consistency and visual quality. Specifically, the Absence of all components causes the Absolute Relative Error (AbsRel) to surge from \textbf{0.434} to \textbf{0.613} and the CD to increase from \textbf{3.334} to \textbf{4.364}, while the PSNR drops from \textbf{22.45} to \textbf{22.24}.

\begin{table}[h!]
    \caption{Ablation study of geometric registration components on the DL3DV dataset. \textbf{SA}: Stereo-Anchor View Selection; \textbf{DA}: Depth-Alignment via RANSAC; \textbf{RC}: Ray-Constrained Gaussian Registration; \textbf{ALL}: All components.}
    \label{tab:ablation}
    \begin{center}
    \begin{tabular}{lcccc}
    \toprule
    Config & PSNR $\uparrow$ & AbsRel $\downarrow$ & CD $\downarrow$ & F-Score $\uparrow$ \\
    \midrule
    w/o SA         & 22.37 & 0.563 & 3.977 & 0.375 \\
    w/o DA         & 22.42 & 0.441 & 3.404 & 0.433 \\
    w/o RC         & 22.45 & 0.448 & 3.491 & 0.436 \\
    w/o DA\&RC     & 22.42 & 0.454 & 3.568 & 0.400 \\
    w/o ALL        & 22.24 & 0.613 & 4.364 & 0.311 \\
    All (Ours)     & \textbf{22.45} & \textbf{0.434} & \textbf{3.334} & \textbf{0.465} \\
    \bottomrule
    \end{tabular}
    \end{center}
    \vspace{-10pt}
\end{table}
\begin{table}[h!]
    \caption{Ablation study of global scene consistency on the DL3DV dataset. \textbf{MV} denotes the Multi-View Gaussian Refinement module.}
    \label{tab:ablation_mv}
    \begin{center}
    \begin{tabular}{lcccc}
    \toprule
    \multirow{2}{*}{Config} & \multicolumn{2}{c}{Target-View} & \multicolumn{2}{c}{Context-View} \\
    \cmidrule(r){2-3} \cmidrule(l){4-5}
    & PSNR $\uparrow$ & SSIM $\uparrow$ & PSNR $\uparrow$ & SSIM $\uparrow$ \\
    \midrule
    DepthSplat & 21.49 & 0.748 & 34.83 & 0.970 \\
    w/o MV     & 22.29 & 0.747 & 34.17 & 0.968 \\
    All (Ours) & \textbf{22.45} & \textbf{0.750} & \textbf{34.98} & \textbf{0.971} \\
    \bottomrule
    \end{tabular}
    \end{center}
\end{table}

\noindent\textbf{w/o Multi-View Gaussian Refinement (MV).} 
Table~\ref{tab:ablation_mv} demonstrates that our multi-view refinement ensures global scene consistency. This module not only enhances target-view performance ($22.29 \rightarrow 22.45$~dB) but also preserves the quality of the original context views ($34.83 \rightarrow 34.98$~dB). Conversely, removing this module (\textit{w/o MV}) results in a dual performance decline: target-view quality drops from our peak performance ($22.45 \rightarrow 22.29$~dB), while context-view quality undergoes a significant collapse relative to the baseline ($34.83 \rightarrow 34.17$~dB). These results confirm that our module effectively prevents catastrophic forgetting.

\subsection{Analysis of Time Efficiency}
\label{subsec:efficiency}
Table~\ref{tab:time_efficiency} details the runtime performance. GSCompleter achieves a total inference time of \textbf{3.16s} on a single NVIDIA H200 GPU. Specifically, while the generative prior (PE-Field) accounts for 1.73s, the subsequent registration stage---comprising alignment, registration, and refinement---is executed in a mere \textbf{1.43s}. As visualized in Fig.~\ref{fig:reggs_compare}, when \textit{excluding} the generative prior to compare registration performance, our core geometric pipeline delivers a \textbf{170$\times$ speedup} over the baseline \textbf{RegGS} ($\sim$4 min). These results underscore that our paradigm is significantly more efficient than traditional registration-based methods.

\begin{table}[htbp]
    \caption{Efficiency analysis of our proposed pipeline. All stages are measured on a single NVIDIA H200 GPU.}
    \label{tab:time_efficiency}
    \begin{center}
    \begin{tabular}{lc}
    \toprule
    Stage & Time Cost (s) \\
    \midrule
    Generative Hypothesis via PE-Field      & 1.73 \\
    Stereo-Anchor View Selection \& 3D Lifting.      & 0.05 \\
    Ray-Constrained Gaussian Registration      & 0.69 \\
    Multi-view Gaussian Integration \& Refinement       & 0.53 \\
    \midrule
    Total                          & 3.17 \\
    \bottomrule
    \end{tabular}
    \end{center}
    \vspace{-10pt}
\end{table}

\subsection{Analysis of Long Sequence Completion} 
\label{subsec:appendix_long_seq_completion}

We evaluate our long-sequence completion performance against RegGS, following their protocol by utilizing ground-truth (graph-truth) images for registration comparison.

\begin{figure}[t!]
    \centering
    \includegraphics[width=1.0\linewidth]{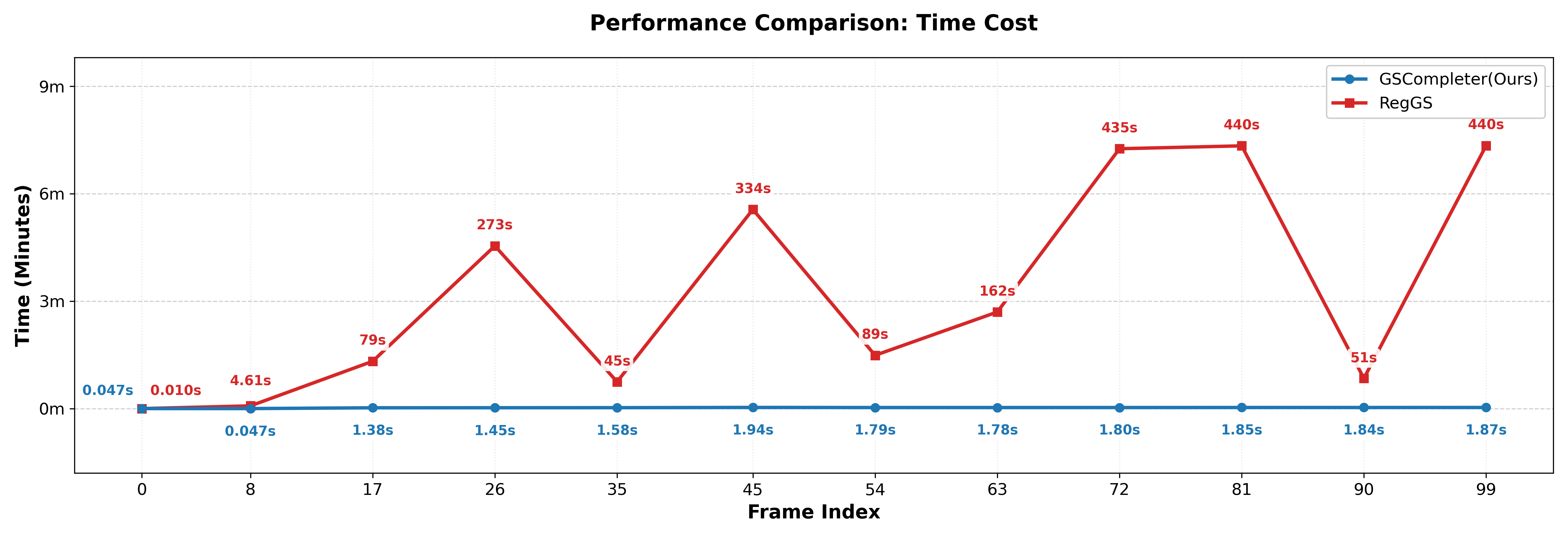}
    \caption{\textbf{Quatitative comparison on long sequences.} RegGS suffers from progressive blurring and structural drift due to metric scale instability. In contrast, GSCompleter maintains sharp details and global consistency, effectively rectifying artifacts in challenging frames.}
    \label{fig:long_seq_comp}
\end{figure}

\begin{figure}[t!]
    \centering
    \includegraphics[width=1.0\linewidth]{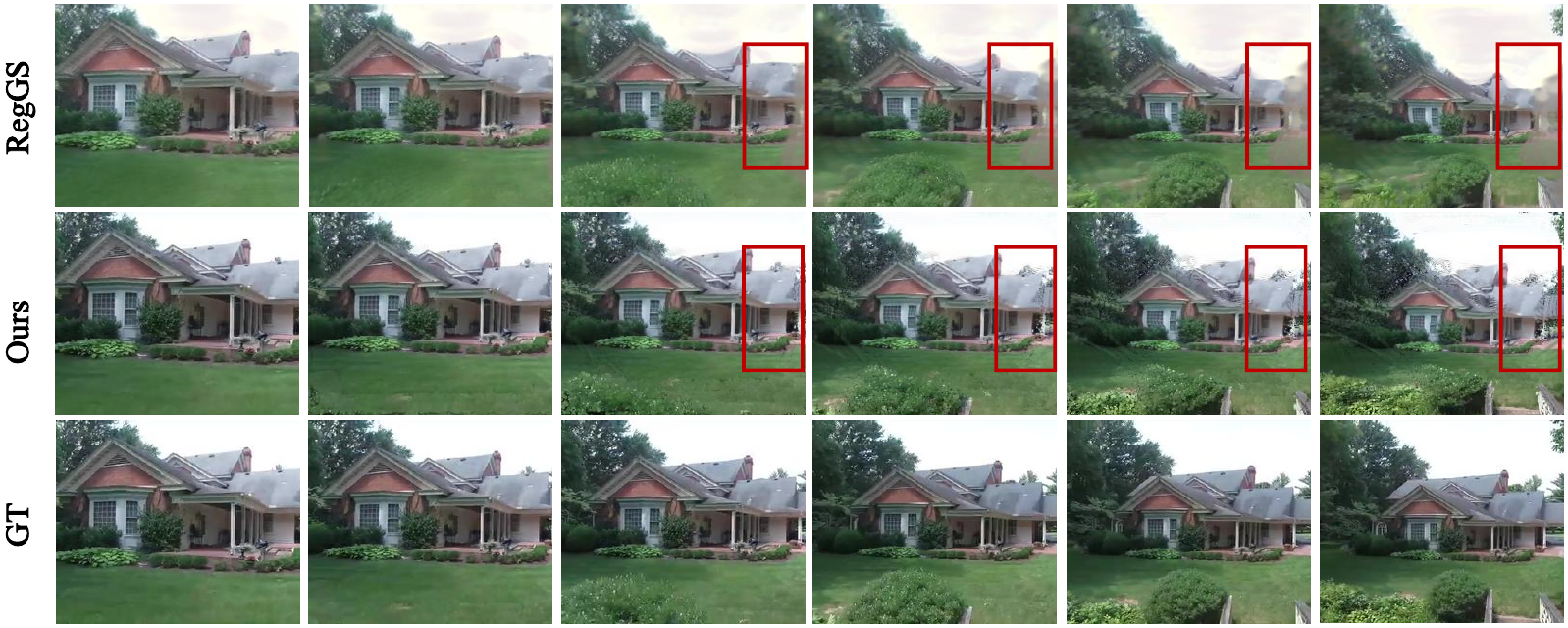}
    \caption{\textbf{Qualitative comparison on long sequences.} To isolate and compare core registration capabilities, we exclude the global refinement stage of RegGS, ensuring a direct baseline comparison. RegGS exhibits artifacts and blur due to error accumulation, while ours preserves high-frequency details and structural integrity.}
    \label{fig:visual_comp}
\end{figure}

\noindent\textbf{Geometric Consistency} 
As visualized in Fig.~\ref{fig:long_seq_comp} and Fig.~\ref{fig:visual_comp}, RegGS suffers from severe temporal error accumulation. Because it relies on iterative optimization across sequential intervals, minor geometric inaccuracies in early stages inevitably propagate and compound. This leads to catastrophic geometric drift, blurred textures, and significant artifacts in later frames. In contrast, our \textit{Stereo-Anchor View Selection} mechanism maintains high structural integrity throughout the trajectory. By anchoring each completion view to a reliable geometric reference, GSCompleter effectively prevents cumulative drift, preserving sharp details even in extended sequences.

\noindent\textbf{Metric Scale Stability} 
A critical failure mode for RegGS is its lack of metric scale consistency over time. As shown in our quantitative analysis, the unstable scale in RegGS causes misaligned geometry to aggregate, resulting in a significantly lower average PSNR of \textbf{19.74 dB}. Conversely, our method leverages a consistent metric scale to ensure robust registration, achieving an average PSNR of \textbf{23.39 dB}. This scale-aware approach allows us to maintain alignment precision that RegGS loses as the sequence length increases.

\noindent\textbf{Computational Efficiency and Reliability} 
Beyond reconstruction quality, our method demonstrates superior efficiency and operational stability. RegGS exhibits extreme variance in computational cost; its optimization time per interval is highly unpredictable, frequently spiking from several seconds to over 440s ($\approx$ 7.3 minutes). In contrast, our approach provides a strictly deterministic time cost, consistently completing each view in approximately \textbf{1.8s}. This represents a speedup of over \textbf{200$\times$} compared to the peak latency of RegGS, ensuring a fast and reliable pipeline for large-scale scene completion.

\subsection{Analysis of 1-DoF vs 3-DoF Registration}
We evaluate the necessity of Ray-Constrained Registration by comparing 1-DoF optimization against unconstrained 3-DoF optimization (Table~\ref{tab:1_dof}). 
The results show that while 3-DoF allows for marginal F-Score gains, it triggers a severe geometric collapse where AbsRel surges by 79\% and structural metrics like CD nearly stagnate. 
In contrast, our 1-DoF constraint locks the optimization to the depth axis, preserving metric stability while significantly reducing CD by 0.1006. 
By suppressing floaters inherent in high-DoF optimization, 1-DoF achieves superior rendering quality (23.93 PSNR) and successfully reconciles global geometric rigor with local visual fidelity.

% \begin{table}[htbp]
%     \centering
%     \setlength{\tabcolsep}{3pt} % 调节列间距（默认通常为 6pt）
%     \caption{Quantitative comparison of 1-DoF and 3-DoF registration strategies.}
%     \label{tab:1_dof}
%     \begin{tabular}{lcccccc}
%     \toprule
%     Method & PSNR $\uparrow$ & AbsRel $\downarrow$ & CD $\downarrow$ & F-Score $\uparrow$  \\
%     \midrule
%     Baseline (Pre-Registration) & -  & 0.2319 & 1.4354 & 0.7154 \\
%     3-DoF (Unconstrained) & 23.28  & 0.4149 & 1.4168 & 0.7216 \\
%     \textbf{1-DoF (Ray-constrained)} & \textbf{23.93} & \textbf{0.2317} & \textbf{1.3348} & \textbf{0.7346} \\
%     \bottomrule
%     \end{tabular}
% \end{table}

\begin{table}[htbp]
    \centering
    \setlength{\tabcolsep}{3pt} % 调节列间距（默认通常为 6pt）
    \caption{Quantitative comparison of 1-DoF and 3-DoF registration strategies.}
    \label{tab:1_dof}
    \begin{tabular}{lcccccc}
    \toprule
    Method & PSNR $\uparrow$ & AbsRel $\downarrow$ & CD $\downarrow$ & F-Score $\uparrow$  \\
    \midrule
    Baseline (Pre-Registration) & -  & 0.4541 & 3.5541 & 0.4028 \\
    3-DoF (Unconstrained) & 22.39  & 0.5139 & 3.5167 & 0.4155 \\
    \textbf{1-DoF (Ray-constrained)} & \textbf{22.45} & \textbf{0.4333} & \textbf{3.3241} & \textbf{0.4656} \\
    \bottomrule
    \end{tabular}
\end{table}

\subsection{Analysis of Depth Alignment Strategy}
\label{sec:exp_alignment_stability}
As shown in Table~\ref{tab:alignment_analysis} and Fig.~\ref{fig:fig_densification}, direct alignment at the target pose is unstable due to unobserved voids. These voids provide insufficient geometric constraints for RANSAC estimation. In contrast, our strategy leverages the anchor view's dense depth to impose spatial constraints. This resolves alignment instabilities and yields a $\textbf{16.8\%}$ F-Score improvement ($0.398 \rightarrow 0.465$), demonstrating the effectiveness of anchor-based guidance.

\begin{table}[htbp]
    \caption{Quantitative comparison of alignment strategies: Target-based depth alignment vs. Anchor-based alignment (Ours).}
    \label{tab:alignment_analysis}
    \setlength{\tabcolsep}{3pt} % 调节列间距（默认通常为 6pt）
    \begin{center}
    \begin{tabular}{lcccc}
    \toprule
    Method & PSNR $\uparrow$ & AbsRel $\downarrow$ & CD $\downarrow$ & F-Score $\uparrow$ \\
    \midrule
    Target-based Alignment & 22.42 & 0.456 & 3.658 & 0.398 \\
    Anchor-based Alignment(\textbf{Ours}) & \textbf{22.45} & \textbf{0.434} & \textbf{3.334} & \textbf{0.465} \\
    \bottomrule
    \end{tabular}
    \end{center}
\end{table}

\section{Discussion} 
\label{sec:discussion}

Experimental results demonstrate that the ``Generate-then-Register'' paradigm our proposed offers significant advantages in execution speed and robustness over the ``Repair-then-Distill'' framework. Notably, by bypassing Gaussian densification, GSCompleter achieves high-speed completion even in the presence of extensive viewpoint voids.

\subsection{Limitations and Future Work} 
\label{subsec:limitation}
Despite these advancements, certain limitations remain. 
(1) Generative Quality: Our framework is upper-bounded by the 2D generative prior. If the PE-Field yields geometrically inconsistent content or blurry textures in unobserved views, noisy pseudo-ground truth may ultimately compromise the final completion outcomes.
(2) Depth Dependency: The pipeline relies heavily on accurate metric depth estimation. Base models that lack reliable metric depth estimation capabilities (e.g., PixelSplat~\cite{charatan2024pixelsplat}) can introduce primitive misalignment, which ultimately degrades the final completion fidelity.

\subsection{Versatility Evaluation on Autonomous Driving Scenarios} 
\label{subsec:ad_completion}
\begin{figure}[ht!]
    \centering
    \includegraphics[width=1.0\linewidth]{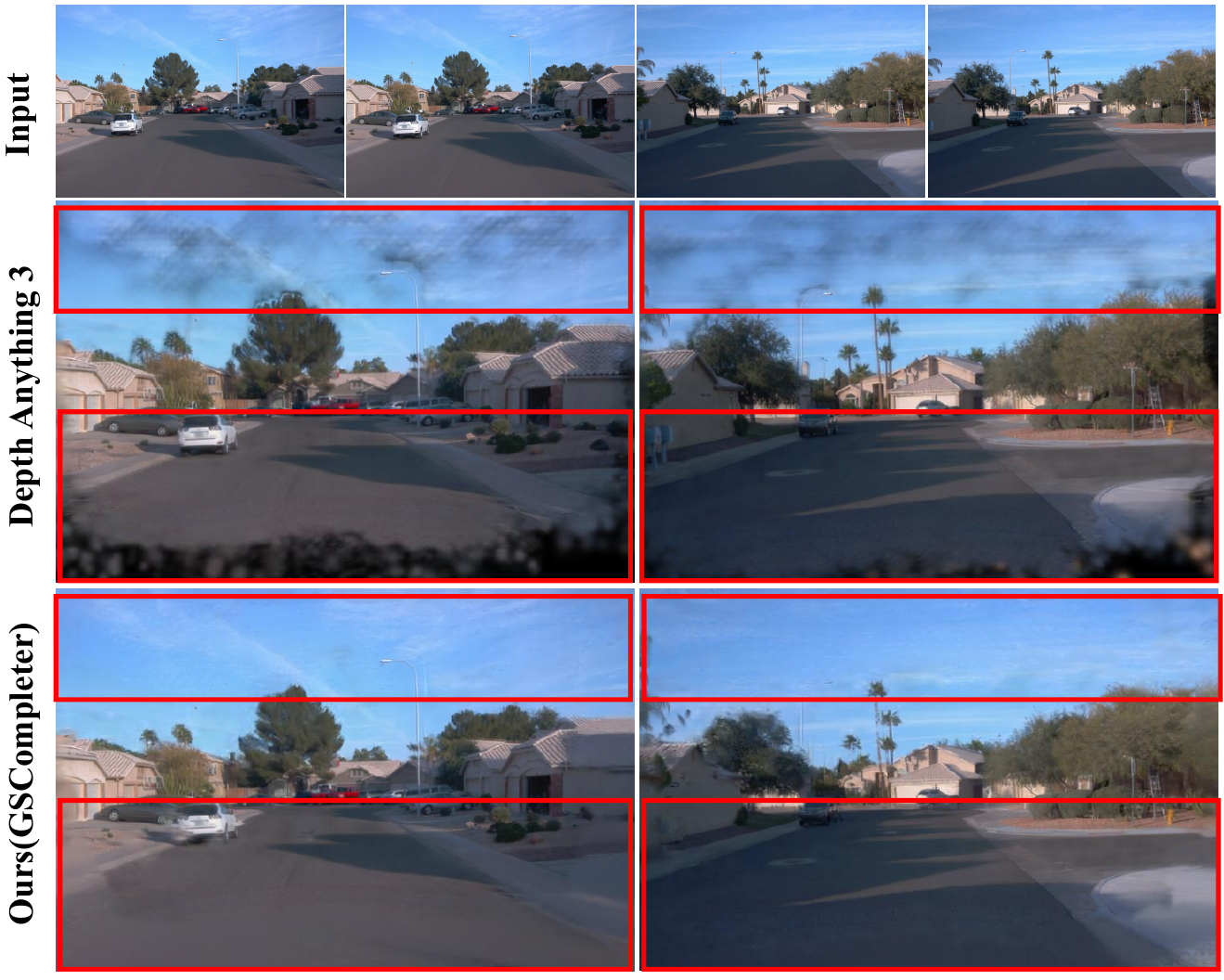}
    \caption{\textbf{Qualitative comparison on autonomous driving (AD) scenes.} By comparing the baseline DA3~\cite{lin2025da3} with ours (DA3 + GSCompleter) on the Waymo Open Dataset, the qualitative results demonstrate the strong versatility and cross-domain generalization of our proposed method.}
    \label{fig:ad_completion}
\end{figure}

We further evaluate the cross-domain performance of GSCompleter in challenging autonomous driving (AD) scenarios. Following the standard evaluation protocol of two context views and one target view in Sec.~\ref{subsec:datasets_evaluation}, we employ Depth Anything 3 (DA3)~\cite{lin2025da3} as the feed-forward baseline to provide the initial Gaussians. As shown in the qualitative results in Fig.~\ref{fig:ad_completion}, our method successfully recovers missing content and produces high-quality scene completions. This consistent performance in AD environments underscores the robust versatility and practical applicability of our plugin across diverse domains.

\begin{figure}[ht!]
    \centering
    \includegraphics[width=1.0\linewidth]{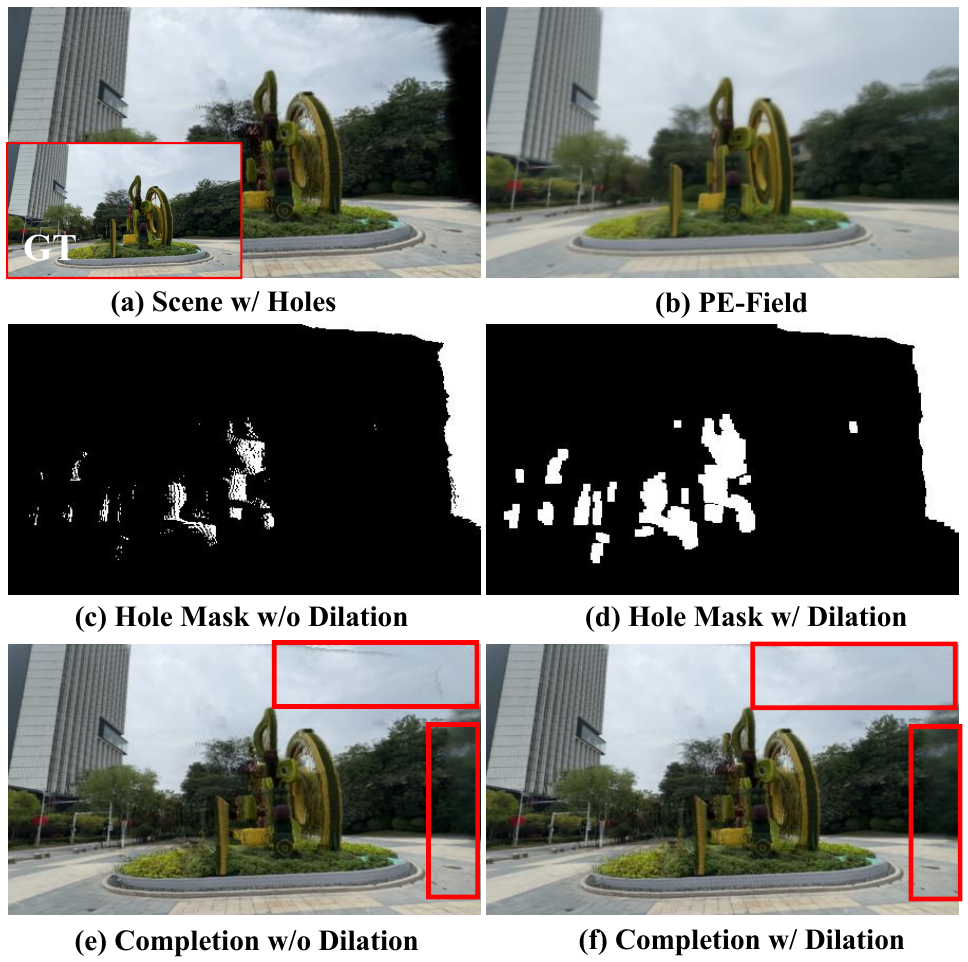}
    \caption{Qualitative comparison of seam addressing via morphological mask expansion. The dilation of the hole mask effectively covers the boundary artifacts, resulting in a more seamless visual integration.}
    \label{fig:visible_seams}
\end{figure}

\subsection{Addressing Visible Seams} 
\label{subsec:visible_seams}

In the qualitative results, discernible boundaries remain between the original context and the filled voids due to the binary hard masks generated via the Gaussian opacity scheme. To mitigate these artifacts, we investigated a morphological dilation strategy (using max\_pool2d with $r=3$) to expand the hole masks and facilitate smoother blending. As shown in Fig.~\ref{fig:visible_seams}, dilation effectively alleviates boundary seams.

\section{Conclusion}

In this paper, we propose GSCompleter, a novel distillation-free plugin that shifts 3DGS scene completion from unstable iterative optimization to a robust ``Generate-then-Register'' workflow. We first leverage a 3D-aware generative prior to synthesize plausible 2D reference images, which are then explicitly lifted into 3D Gaussian primitives with precise scale via a robust Stereo-Anchor View Selection mechanism. To ensure seamless integration, we introduce a Ray-Constrained Registration strategy that restricts new Gaussian positions along camera rays, achieving rapid alignment while mitigating texture drifting. Extensive experiments across three benchmarks and various baseline categories—including feed-forward, optimization-based, and registration-based methods—demonstrate that our approach significantly enhances rendering performance. Furthermore, its superior results in both general indoor/outdoor environments and autonomous driving scenarios fully validate its strong versatility and state-of-the-art (SOTA) performance.

\iffalse
{\appendix[Proof of the Zonklar Equations]
Use $\backslash${\tt{appendix}} if you have a single appendix:
Do not use $\backslash${\tt{section}} anymore after $\backslash${\tt{appendix}}, only $\backslash${\tt{section*}}.
If you have multiple appendixes use $\backslash${\tt{appendices}} then use $\backslash${\tt{section}} to start each appendix.
You must declare a $\backslash${\tt{section}} before using any $\backslash${\tt{subsection}} or using $\backslash${\tt{label}} ($\backslash${\tt{appendices}} by itself
 starts a section numbered zero.)}
\fi

%\begin{thebibliography}{1}
\bibliographystyle{IEEEtran}

\bibliography{gscompleter}

%\end{thebibliography}

\vspace{-50pt}

\begin{IEEEbiography}[{\includegraphics[width=1in,height=1.25in,clip,keepaspectratio]{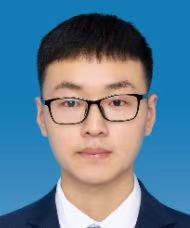}}]{Ao Gao} is current pursuing the Ph.D. degree at the School of Computer Science and Technology, East China Normal University. Before this, he received his Master’s degree from Donghua University, China in 2023. His research interests focus on 3D Reconstruction and Neural Rendering.
\end{IEEEbiography}

\vspace{-50pt}

\begin{IEEEbiography}[{\includegraphics[width=1.0in,height=1.25in,clip,keepaspectratio]{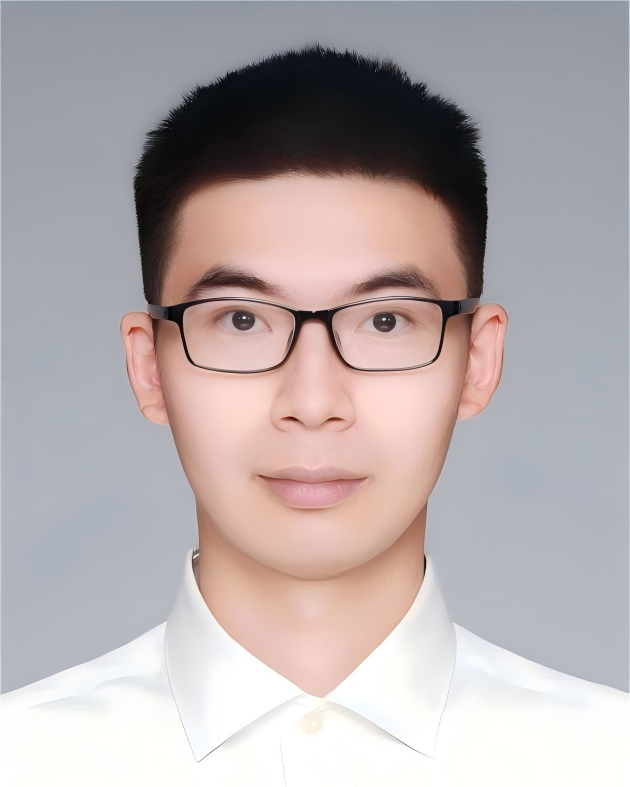}}]{Jingyu Gong} received his Ph.D. degrees from the Department of Computer Science and Technologies, Shanghai Jiao Tong University. He is now an Associate Research Professor with the School of Computer Science and Technology, East China Normal University, China. His research interests include computer 3D vision and computer graphics.
\end{IEEEbiography}

\vspace{-55pt}

\begin{IEEEbiography}[{\includegraphics[width=1in,height=1.25in,clip,keepaspectratio]{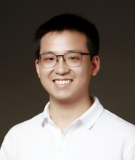}}]{Xin Tan} received his Ph.D. degree in Computer Science from Shanghai Jiao Tong University in 2022. He received his B.Eng. degree in Automation from Chongqing University, China in 2017. He is currently the Research Professor at the School of Computer Science and Technology, East China Normal University, China. His research interests lie in computer vision and deep learning. He serves as a program committee member/reviewer for CVPR, ICCV, ECCV, AAAI, IJCAI, IEEE TPAMI, TIP and IJCV.
\end{IEEEbiography}

\vspace{-45pt}

\begin{IEEEbiography}[{\includegraphics[width=1in,height=1.25in,clip,keepaspectratio]{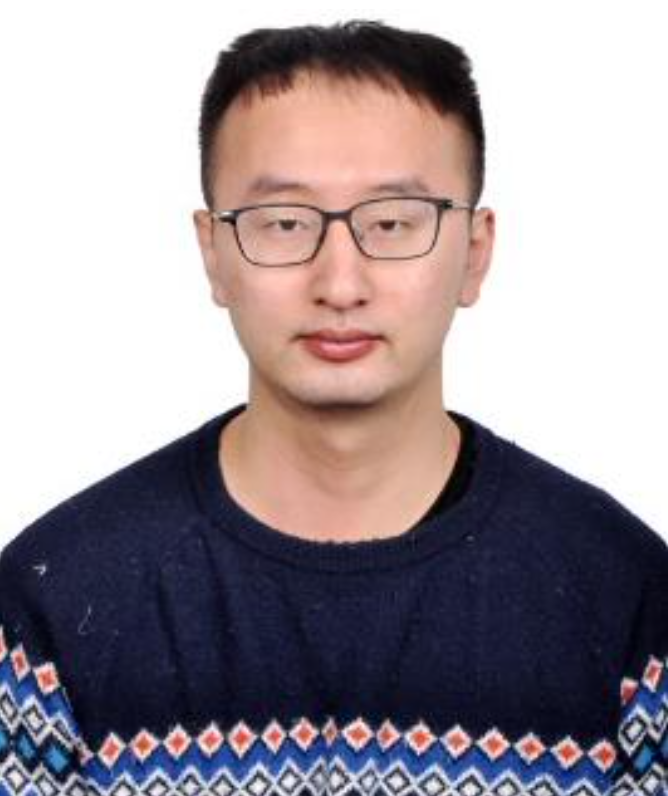}}]{Zhizhong Zhang} received the Ph.D. degree in pattern recognition and intelligent systems from the Institute of Automation, Chinese Academy of Sciences (CAS), in 2020. He is currently an Associate Professor with the School of Computer Science and Technology, East China Normal University. His research interests include image processing, computer vision, machine learning, and pattern recognition.
\end{IEEEbiography}

\vspace{-55pt}

\begin{IEEEbiography}[{\includegraphics[width=1in,height=1.25in,clip,keepaspectratio]{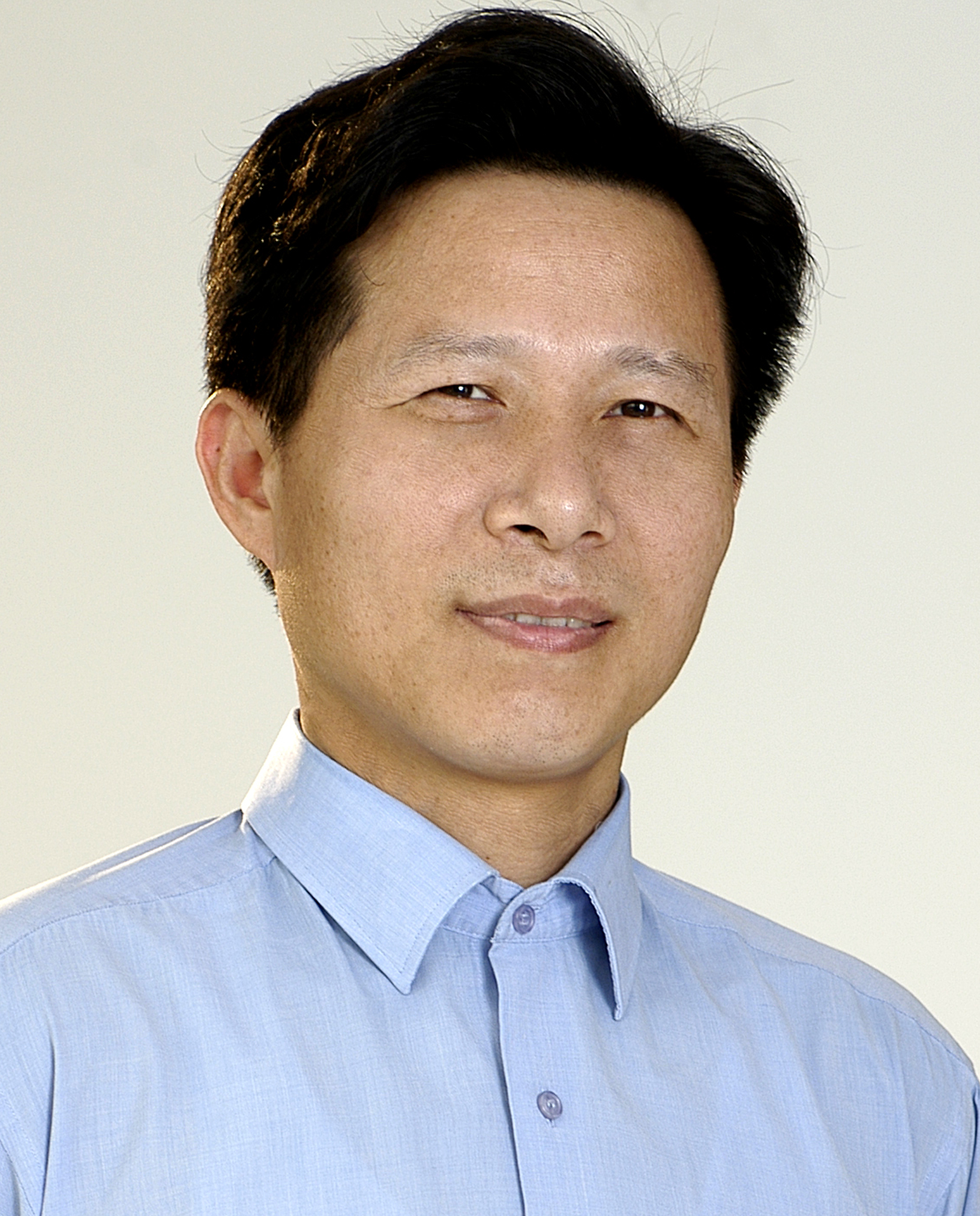}}]{Lizhuang Ma} received his B.S. and Ph.D. de- grees from the Zhejiang University, China in 1985 and 1991, respectively. He is now a Distinguished Professor, at the Department of Computer Science and Engineering, Shanghai Jiao Tong University, China and the School of Computer Science and Technology, East China Normal University, China. He was a Visiting Professor at the Frounhofer IGD, Darmstadt, Germany in 1998, and a Visiting Professor at the Center for Advanced Media Technology, Nanyang Technological University, Singapore from 1999 to 2000. His research interests include computer vision, computer aided geometric design, computer graphics, scientific data visualization, computer animation, digital media technology, and theory and applications for computer graphics, CAD/CAM. He serves as the reviewer of IEEE TPAMI, IEEE TIP, IEEE TMM, CVPR, AAAI etc.
\end{IEEEbiography}

\vspace{-45pt}

\begin{IEEEbiography}[{\includegraphics[width=1in,height=1.25in,clip,keepaspectratio]{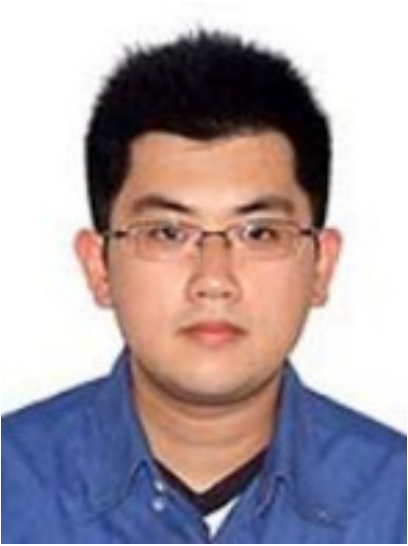}}]{Yuan Xie} received the PhD degree in Pattern Recognition and Intelligent Systems from the Institute of Automation, Chinese Academy of Sciences (CAS), in 2013. He is currently a full professor with the School of Computer Science and Technology, East China Normal University, Shanghai, China. His research interests include image processing, computer vision, machine learning, and pattern recognition. He has published around 90 papers in major international journals and conferences including the IJCV, IEEE TPAMI, TIP, TNNLS, TCYB, NIPS, ICML, CVPR, ECCV, ICCV, etc. He also has served as a reviewer for more than 15 journals and conferences. Dr. Xie received the National Science Fund for Excellent Young Scholars 2022.
\end{IEEEbiography}

% \vfill

% {\appendices
% \section*{Proof of the First Zonklar Equation}
% Appendix one text goes here.
% You can choose not to have a title for an appendix if you want by leaving the argument blank
% \section*{Proof of the Second Zonklar Equation}
% Appendix two text goes here.}

\section{Appendix}

\subsection{Ablation study of Large-Angle Completion Robustness}
\label{subsec:appendix_exp_span_robustness}
To evaluate the robustness of our framework under extreme geometric variations, we conduct a stress test on 140 test scenes from the DL3DV dataset. This experiment follows the $n$-$k$ evaluation protocol established in Sec.~\ref{subsec:datasets_evaluation}.

\noindent\textbf{Regular Scenarios.} As summarized in Table~\ref{tab:extrapolation_span}, the baseline distance between context views expands as the frame interval $k$ increases, leading to a drastic reduction in visual overlap and intensifying the challenge for both geometric inference and generative completion. While the performance of the feed-forward baseline (DepthSplat) degrades sharply with increasing spans, GSCompleter exhibits remarkable robustness. Our method consistently outperforms the baseline across all configurations, with the performance margin widening as viewpoint variations escalate. Notably, in the $n$-$30$ scenario, our paradigm achieves a PSNR of 18.33 dB, surpassing the baseline by \textbf{+1.31 dB}. Furthermore, our approach demonstrates superior perceptual quality (lower FID) across all intervals.

\noindent\textbf{Extreme Scenarios.} We employ an extreme extrapolation setting by selecting context views at indices 0 and 30, while designating the target view at index 99. In this configuration, the spatial overlap in the majority of scenes is near-zero, posing a significant challenge to geometric consistency. As shown in Table~\ref{tab:stress_test}, GSCompleter maintains a substantial performance lead over the baseline even in this extremely sparse configuration, achieving a PSNR gain of \textbf{2.52 dB} (13.31 vs. 10.79). Notably, while the rendering metrics (PSNR/SSIM) remain competitive, the introduction of the Stereo-Anchor (SA) view selection mechanism yields consistent improvements in geometric accuracy, as evidenced by the reduction in Absolute Relative Error (AbsRel) and Chamfer Distance (CD). These findings demonstrate that our system effectively leverages metric priors to ensure structural integrity, even when spatial overlap is minimal.

\begin{table}[h!]
    \caption{\textbf{Robustness analysis against varying extrapolation spans.} We evaluate performance on the DL3DV dataset with increasing frame intervals ($k$). A larger $k$ denotes sparser input sampling, leading to aggravated geometric difficulty. GSCompleter consistently maintains superior PSNR and perceptual quality, highlighting its effectiveness in handling large-baseline inputs.}
    \label{tab:extrapolation_span}
    \centering
    \small % 可选：缩小字号使表格更紧凑
    \setlength{\tabcolsep}{3pt} % 调节列间距（默认通常为 6pt）
    \begin{tabular}{lcccccc}
    \toprule
    & \multicolumn{2}{c}{\textbf{n-10}} & \multicolumn{2}{c}{\textbf{n-20}} & \multicolumn{2}{c}{\textbf{n-30}}\\
    \cmidrule(lr){2-3} \cmidrule(lr){4-5} \cmidrule(lr){6-7}
    Method & PSNR $\uparrow$ & FID $\downarrow$ & PSNR $\uparrow$ & FID $\downarrow$ & PSNR $\uparrow$ & FID $\downarrow$ \\
    \midrule
    DepthSplat & 21.49 & 65.21 & 18.13 & 104.15 & 17.02 & 124.80 \\
    \textbf{Ours} & \textbf{22.45} & \textbf{60.95} & \textbf{19.44} & \textbf{99.22} & \textbf{18.33} & \textbf{122.12} \\
    \bottomrule
    \end{tabular}
\end{table}

% \begin{table}[h!] % 通常顶会建议用 [t] 或 [b]
%     \centering
%     \caption{\textbf{Quantitative Evaluation of Stress Test on DL3DV.} Under extreme extrapolation settings, GSCompleter significantly outperforms the baseline. Note that the Stereo-Anchor Selection mechanism primarily enhances geometric fidelity (AbsRel, CD, and F-Score).}
%     \label{tab:stress_test}
%     \begin{tabular}{lcccccc}
%         \toprule
%         Method & PSNR$\uparrow$ & SSIM$\uparrow$ & LPIPS$\downarrow$ & AbsRel$\downarrow$ & CD$\downarrow$ & F-Score$\uparrow$ \\ 
%         \midrule
%         DepthSplat~\cite{xu2025depthsplat} & 10.79 & 0.294 & 0.605 & -- & -- & -- \\
%         Ours (w/o SA Selection) & \textbf{13.31} & 0.318 & \textbf{0.603} & 0.703 & 7.172 & 0.291 \\
%         \textbf{Ours (GSCompleter)} & \textbf{13.31} & \textbf{0.319} & \textbf{0.603} & \textbf{0.660} & \textbf{7.094} & \textbf{0.297} \\ 
%         \bottomrule
%     \end{tabular}
% \end{table}

\begin{table}[ht!]
    \centering
    \small % 可选：缩小字号使表格更紧凑
    \caption{\textbf{Quantitative Evaluation of Stress Test on DL3DV.} Under extreme extrapolation settings, GSCompleter significantly outperforms the baseline. Note that the Stereo-Anchor Selection mechanism primarily enhances geometric fidelity (AbsRel, CD, and F-Score).}
    \label{tab:stress_test}
    \setlength{\tabcolsep}{2pt} % 调节列间距（默认通常为 6pt）
    \begin{tabular}{lccccc}
        \toprule
        Method & PSNR$\uparrow$ & SSIM$\uparrow$ & AbsRel$\downarrow$ & CD$\downarrow$ & F-Score$\uparrow$ \\ 
        \midrule
        DepthSplat~\cite{xu2025depthsplat} & 10.79 & 0.294 & -- & -- & -- \\
        Ours (w/o SA) & \textbf{13.31} & 0.318 & 0.703 & 7.172 & 0.291 \\
        \textbf{Ours (GSCompleter)} & \textbf{13.31} & \textbf{0.319} & \textbf{0.660} & \textbf{7.094} & \textbf{0.297} \\ 
        \bottomrule
    \end{tabular}
\end{table}

% \subsection{Ablation study of PE-Field Prior Quality}
% \label{sec:exp_pe_field}

% As illustrated in Table \ref{tab:pe_field_analysis}, Step 4 emerges as the optimal configuration, achieving peak performance in geometric reconstruction with a minimal AbsRel of 0.433 and Chamfer Distance of 3.324. While fewer steps lead to under-optimization and extending to Step 8 induces marginal geometric degradation, Step 4 strikes a superior balance between perceptual fidelity and structural precision. Specifically, it significantly enhances LPIPS and FID over the baseline while maintaining robust 3D structural integrity.

% \input{tables/pe_field}

\subsection{Ablation Study on PE-Field Inference Steps}
\label{sec:exp_pe_field}

As illustrated in Table \ref{tab:pe_field_analysis}, we evaluate the sensitivity of the PE-Field prior quality by varying the number of inference steps. The 4-step configuration emerges as the optimal choice, striking a superior balance between efficiency and reconstruction accuracy. Specifically, it achieves the peak performance in geometry with a minimal AbsRel of 0.433 and Chamfer Distance (CD) of 3.324, while maintaining an excellent LPIPS score of 0.182. Although reducing the steps to 2 yields slightly faster inference (1.03s), such sub-optimal image generation leads to noticeable geometric degradation (CD increases to 3.392). Conversely, further extending the inference to 8 or 16 steps offers no perceptual gains (LPIPS plateaus at 0.182), while doubling or quadrupling the computational time cost. Consequently, 4 steps serve as the default setting to guarantee robust structural precision without compromising efficiency.

\begin{table}[htbp]
    \caption{Sensitivity analysis of the PE-Field inference steps for prior quality.}
    \label{tab:pe_field_analysis}
    \setlength{\tabcolsep}{3pt} % 调节列间距（默认通常为 6pt）
    \begin{center}
    \begin{tabular}{lccccc}
    \toprule
    Configuration & PSNR $\uparrow$ & LPIPS $\downarrow$ & AbsRel $\downarrow$ & CD $\downarrow$ & Time Cost (s) $\downarrow$ \\
    \midrule
    Step 2  & \textbf{22.52} & 0.183 & 0.437 & 3.392 & \textbf{1.03} \\
    Step 4  & 22.45 & \textbf{0.182} & \textbf{0.433} & \textbf{3.324} & 1.83 \\
    Step 8  & 22.38 & \textbf{0.182} & 0.441 & 3.349 & 3.43 \\
    Step 16 & 22.34 & \textbf{0.182} & 0.443 & 3.329 & 6.62 \\
    \bottomrule
    \end{tabular}
    \end{center}
\end{table}

\subsection{Ablation study of Stereo-Anchor Rotation Angle Constraint}
\label{sec:exp_pe_field}

Table~\ref{tab:sa_fallback} investigates the sensitivity of system performance to the rotation threshold $\theta_{max}$. While a restrictive $30^\circ$ constraint triggers the fallback mechanism (FB=1) and yields suboptimal scores, a $45^\circ$ threshold eliminates fallbacks and achieves peak performance (e.g., 22.452 PSNR and 3.324 CD). Given that results plateau beyond this point, we adopt $45^\circ$ as the default to ensure optimal selection flexibility without compromising stability.

\begin{table}[ht!]
    \centering
    \small 
    \caption{Sensitivity analysis of SA rotation constraints under 140 test sets of DL3DV dataset. We select $n=10$ as the experimental setting. \textbf{FB} denotes the Fallback count.}
    \label{tab:sa_fallback}
    \setlength{\tabcolsep}{3pt} 
    \begin{tabular}{ccccccc}
        \toprule
        Threshold & FB$\downarrow$ & PSNR$\uparrow$ & SSIM$\uparrow$ & AbsRel$\downarrow$ & CD$\downarrow$ & F-Score$\uparrow$ \\ 
        \midrule
        $30^\circ$ & 1 & 22.450 & 0.7495 & 0.434 & 3.331 & 0.465 \\
        $45^\circ$ & \textbf{0} & \textbf{22.452} & \textbf{0.7496} & \textbf{0.433} & \textbf{3.324} & \textbf{0.466} \\
        $90^\circ$ & \textbf{0} & \textbf{22.452} & \textbf{0.7496} & \textbf{0.433} & \textbf{3.324} & \textbf{0.466} \\ 
        \bottomrule
    \end{tabular}
\end{table}

\subsection{Additional Qualitative Results} 
\label{subsec:more_results}
We provide additional qualitative results on the following pages.

\begin{figure*}[h!]
\centering
\includegraphics[width=1.0\linewidth]{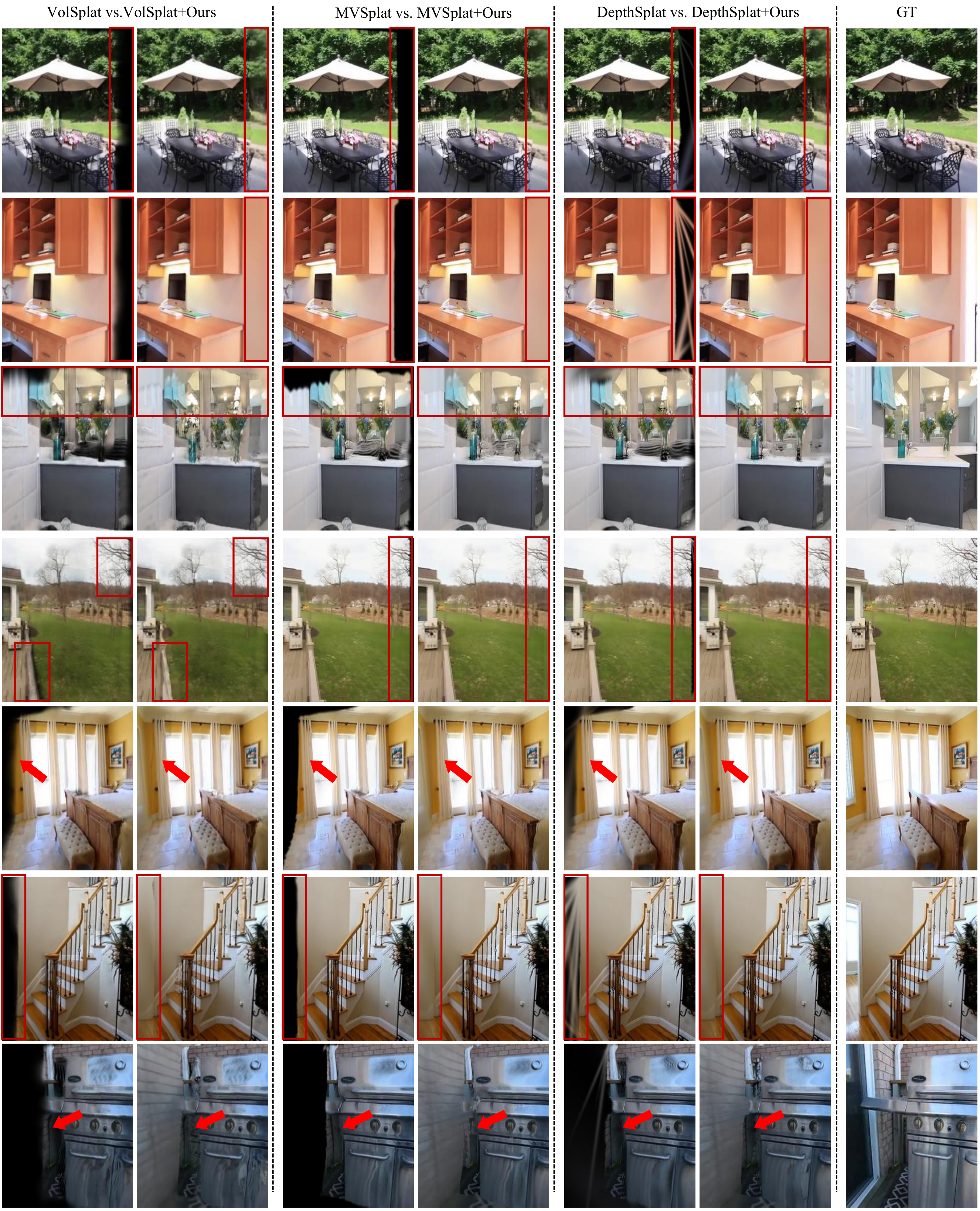}
\caption{More results on the RealEstate10K dataset.}
\label{fig:more_re10k} % Fixed duplicate label
\end{figure*}

\begin{figure*}[h!]
\centering
\includegraphics[width=1.0\linewidth]{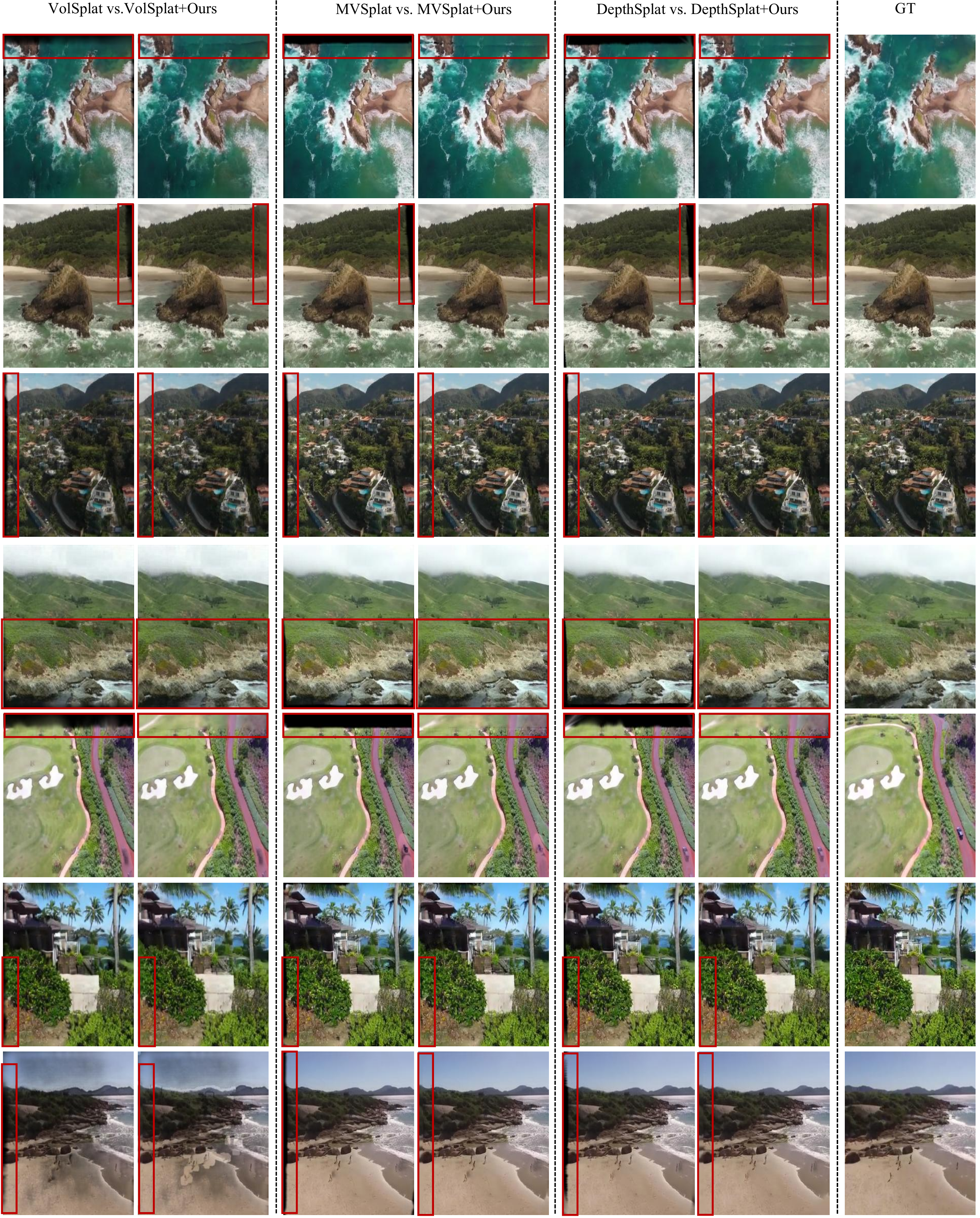}
\caption{More results on the ACID dataset.}
\label{fig:more_acid} % Fixed duplicate label
\end{figure*}

\begin{figure*}[h!]
\centering
\includegraphics[width=1.0\linewidth]{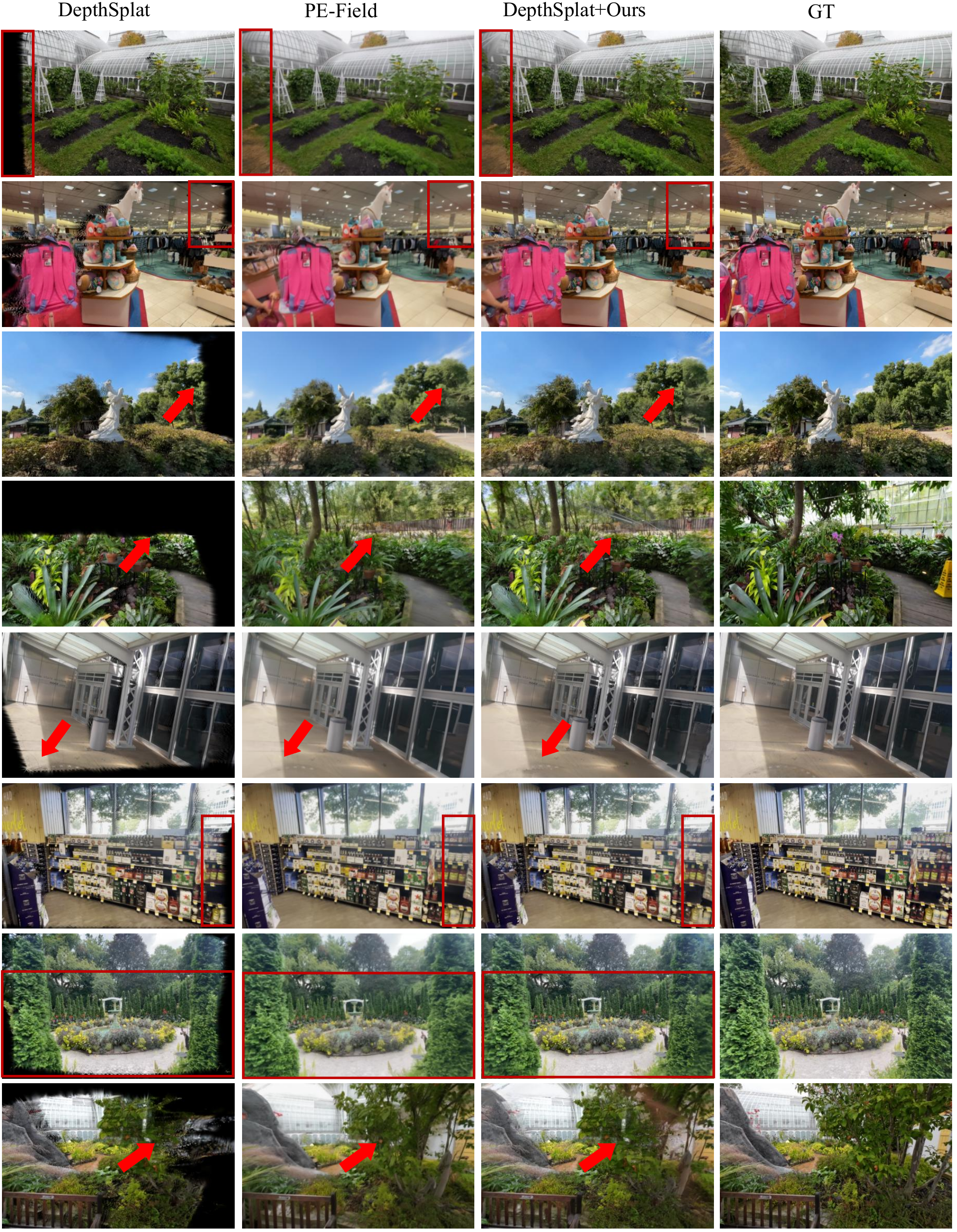}
\caption{More results on the DL3DV dataset.}
\label{fig:more_dl3dv} % Fixed duplicate label
\end{figure*}

\end{document}